\documentclass{article}

\usepackage{microtype}
\usepackage{graphicx}
\usepackage{subfigure}
\usepackage{booktabs} 
\usepackage{natbib}

\usepackage{hyperref}
\usepackage{mathtools}

\DeclarePairedDelimiterX{\infdivx}[2]{(}{)}{%
  #1\;\delimsize\|\;#2%
}

\usepackage[accepted]{icml2024}
\usepackage{amsmath}
\usepackage{amssymb}
\usepackage{mathtools}
\usepackage{amsthm}
\usepackage{siunitx}
\usepackage{bm}

\usepackage[capitalize,noabbrev]{cleveref}

\theoremstyle{plain}

\theoremstyle{definition}

\theoremstyle{remark}

\renewcommand{\vec}[1]{\bm{#1}}

\newcommand{\state}{\vec{x}}

\newcommand{\dataset}{\mathcal{D}}
\newcommand{\lowerbound}{\state_{\mathsf{L}}}
\newcommand{\upperbound}{\state_{\mathsf{U}}}
\newcommand{\given}{\mid}

\DeclarePairedDelimiterXPP\kl[2]{D_{KL}}(){}{#1 \| #2}

\newcommand{\expect}{\mathbb{E}}
\DeclarePairedDelimiterXPP\fullexpect[3]{\expect_{#1 \sim #2}}[]{}{#3}

\newcommand{\pdist}[1]{p(#1)}

\DeclarePairedDelimiterXPP\condqdist[2]{q}(){}{#1 \given #2}
\DeclarePairedDelimiterXPP\optimalcondqdist[2]{q^\star}(){}{#1 \given #2}
\DeclarePairedDelimiterXPP\condpdist[2]{p}(){}{#1 \given #2}
\DeclarePairedDelimiterXPP\condparampdist[3]{p_{#1}}(){}{#2 \given #3}
\DeclarePairedDelimiterXPP\gauss[2]{\mathcal{N}}(){}{#1, #2}

\newcommand{\cdf}{\Phi}
\newcommand{\pdf}{\varphi}

\newcommand{\scale}{\vec{s}}

\newcommand{\identity}{\mathbf{I}}

\newcommand{\drawn}{\sim}

\usepackage[textsize=tiny]{todonotes}

\icmltitlerunning{Diffusion models for large-scale sea-ice modelling}

\begin{document}

\twocolumn[
\icmltitle{Towards diffusion models for large-scale sea-ice modelling}



\icmlsetsymbol{equal}{*}

\begin{icmlauthorlist}
    \icmlauthor{Tobias Sebastian Finn}{enpc}
    \icmlauthor{Charlotte Durand}{enpc}
    \icmlauthor{Alban Farchi}{enpc}
    \icmlauthor{Marc Bocquet}{enpc}
    \icmlauthor{Julien Brajard}{nersc}
\end{icmlauthorlist}

\icmlaffiliation{enpc}{CEREA, Ecole des Ponts and EDF R\&D, Ile-de-France, France}
\icmlaffiliation{nersc}{Nansen Environmental and Remote Sensing Center, Bergen, Norway}

\icmlcorrespondingauthor{Tobias Sebastian Finn}{tobias.finn@enpc.fr}

\icmlkeywords{Generative diffusion, Earth system modelling, sea-ice simulations}

\vskip 0.3in
]



\printAffiliationsAndNotice{} 

\begin{abstract}
We make the first steps towards diffusion models for unconditional generation of multivariate and Arctic-wide sea-ice states.
While targeting to reduce the computational costs by diffusion in latent space, latent diffusion models also offer the possibility to integrate physical knowledge into the generation process.
We tailor latent diffusion models to sea-ice physics with a censored Gaussian distribution in data space to generate data that follows the physical bounds of the modelled variables.
Our latent diffusion models reach similar scores as the diffusion model trained in data space, but they smooth the generated fields as caused by the latent mapping.
While enforcing physical bounds cannot reduce the smoothing, it improves the representation of the marginal ice zone.
Therefore, for large-scale Earth system modelling, latent diffusion models can have many advantages compared to diffusion in data space if the significant barrier of smoothing can be resolved.
\end{abstract}

\section{Introduction}

Generative diffusion \cite{sohl-dickstein_deep_2015, ho_denoising_2020, song_maximum_2021} has revolutionised data generation in computer vision \cite{dhariwal_diffusion_2021, saharia_photorealistic_2022} and other domains \cite{bar-tal_lumiere_2024, huang_noise2music_2023}.
Initial applications to geophysical systems show promise for prediction \cite{leinonen_latent_2023, li_seeds_2023,price_gencast_2024,finn_generative_2024} and downscaling \cite{mardani_generative_2023, wan_debias_2023, brajard_super-resolution_2024}.
To cut costs, data can be mapped into a latent space using a pre-trained autoencoder, where diffusion models can be learned \cite{sinha_d2c_2021, vahdat_score-based_2021, rombach_high-resolution_2022}.
However, the mapping can reduce the model's ability to generate fine-grained data \cite{dai_emu_2023}.

\begin{figure*}[ht]
\begin{center}
\centerline{\includegraphics[width=0.9\textwidth]{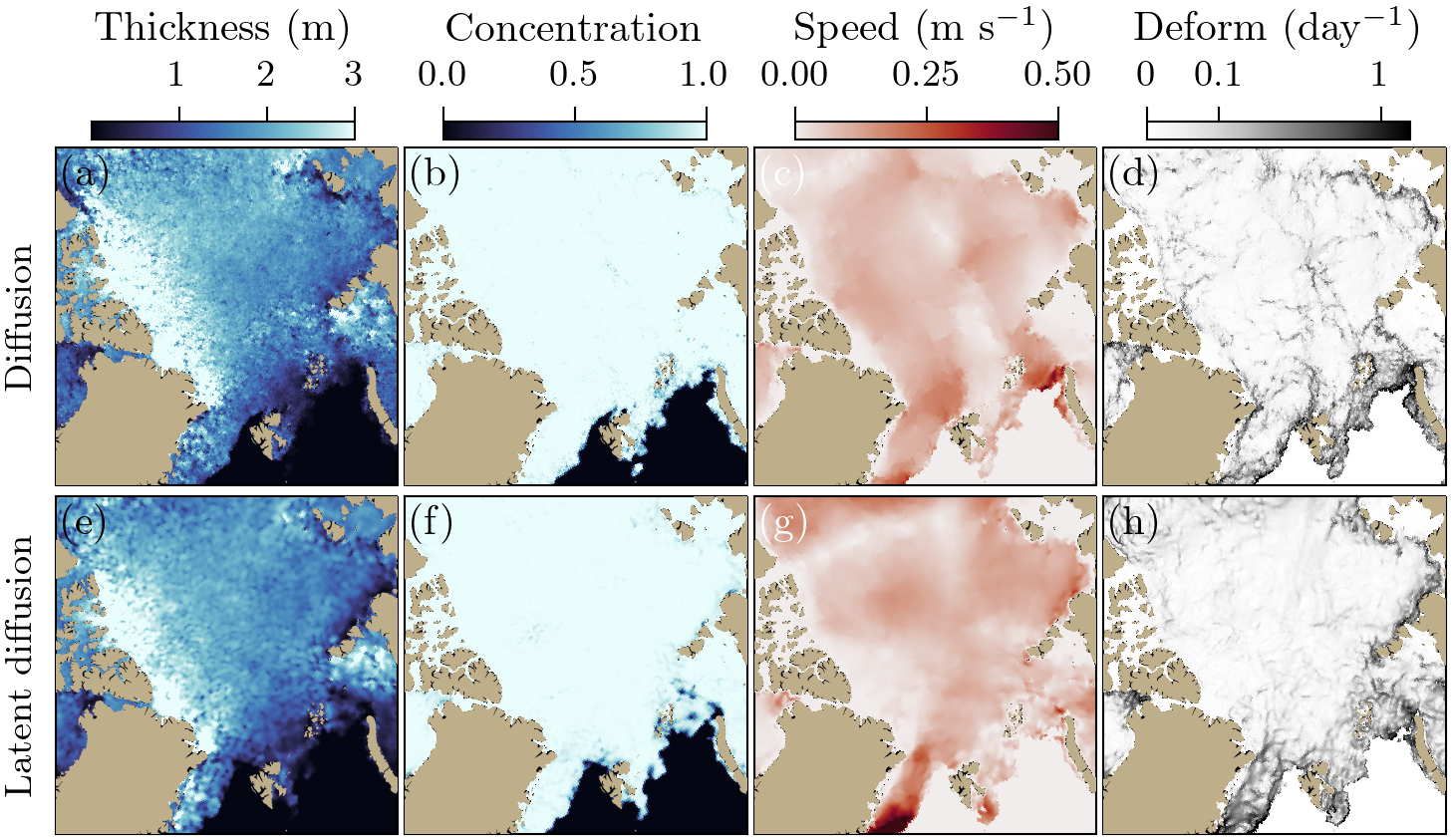}}
\caption{
    Samples generated with a diffusion model in data space (a--d) and a latent diffusion model with censoring (e--h).
    The thickness (a, e) and concentration (b, f) are directly generated, while the speed (c, g) and deformation (d, h) are derived from the velocities.
    Note, for visualisation purpose, only the central Arctic is shown while the whole Arctic is modelled.
    The remaining noise might be caused by using only 20 integration steps with a Heun sampler.
}
\label{fig:samples}
\end{center}
\vskip -0.3in
\end{figure*}

Here, we examine how well latent diffusion models (LDMs) can generate geophysical data compared to diffusion directly applied in data space.
As a proof-of-concept, we unconditionally generate Arctic-wide sea-ice states as simulated with the state-of-the-art sea-ice model neXtSIM \cite{rampal_nextsim_2016, olason_new_2022} at a $\frac{1}{4}^{\circ}$ resolution \cite{boutin_arctic_2023}.
Sea ice, with its discrete elements, anisotropy, and multifractality, poses a challenging problem for generative models and serves as ideal candidate for evaluating LDMs.
We show that while the general structure remains the same, LDMs produces smoother solutions than diffusion models in data space, as presented in Fig. \ref{fig:samples}.

Although generative diffusion is fundamentally defined by its Gaussian diffusion process, we can integrate prior physical information into the encoder and decoder of an LDM, as similarly done in \citet{shmakov_end--end_2023}.
Hence, we tailor the approach of LDMs to Earth system modelling by incorporating physical bounds into the autoencoder.
To bound the variables, we replace the common Gaussian reconstruction loss (mean-squared error) by the log-likelihood of censored Gaussian distributions, which combines regression and classification, enabling us to explicitly account for the bounds of the output during reconstruction.
Such censored distributions improve the representation of the sea-ice extent and thereby enhance the physical consistency of the output, as demonstrated in this study.

\section{Methods}\label{sec:methods}

In the following, we introduce a brief overview on LDMs.
Specifically, we focus on how to make use of prior physical knowledge into such models.
For a more detailed treatment, we refer to Appendix \ref{app:add_methods}.

Our goal is to train neural networks (NNs) to generate state samples ${\widehat{\mathbf{x}} \in \mathbb{R}^{5\times512\times512}}$ that should look like samples drawn from the here-used sea-ice simulation $\mathbf{x} \sim \mathcal{D}$; modelled are the thickness (SIT), concentration (SIC), damage (SID), and the velocities in $x$- (SIU) and $y$-direction (SIV).
For the generation, we employ a diffusion model that works in a lower-dimensional latent space ${\mathbf{z}_{x} \in \mathbb{R}^{16\times 64 \times 64}}$ as spanned by an autoencoder. 
The encoder $\mathrm{enc}(\mathbf{x})$ maps from data space into latent space, while the decoder $\mathrm{dec}(\mathbf{z}_{x})$ maps back into data space.
The encoder and decoder are implemented as fully convolutional NNs (CNNs) with their weights and biases $\boldsymbol{\theta}$ and further described in Appendix \ref{app:ae_architecture}.

The autoencoder should reduce the data dimensionality without compressing the semantic \cite{rombach_high-resolution_2022}.
We nevertheless want to achieve a continuous latent space, from which the decoder can map similar values to similar data points.
Consequently, we apply a variational autoencoder \cite{kingma_auto-encoding_2013}, where we downweight the Kullback-Leibler divergence by $\beta = 0.001$ \cite{higgins_beta-vae:_2016}.
Given a data sample $\mathbf{x}$, the loss function for the autoencoder then reads,
\begin{subequations}
    \begin{align}
        \mathcal{L}(\mathbf{x}, \bm{\theta}) = &-\expect_{q(\mathbf{z}_{x} \given \mathbf{x}, \bm{\theta})}\Big[\log\big(p(\mathbf{x} \given \mathbf{z}_{x}, \bm{\theta})\big)\Big]\label{eq:recon_loss}\\
        &+ \beta \kl[\Big]{q(\mathbf{z}_{x} \given \mathbf{x}, \bm{\theta})}{p(\mathbf{z}_{x})},\label{eq:prior_loss}
    \end{align}\label{eq:autoencoder_loss}
\end{subequations}
which is minimised by averaging the loss across a mini-batch of samples and using a variant of gradient descent.

The encoder maps a data sample into the mean and standard deviation of an assumed Gaussian distribution in latent space, $q(\mathbf{z}_{x} \given \mathbf{x}, \bm{\theta}) = \mathcal{N}\left(\mu_{\mathsf{enc}, \bm{\theta}}(\mathbf{x}), (\sigma_{\mathsf{enc}, \bm{\theta}}(\mathbf{x}))^2 \identity\right)$, while we assume a prior Gaussian distribution with mean $\bm{0}$ and a diagonal covariance $\mathbf{I}$, $p(\mathbf{z}_{x})=\mathcal{N}\left(\bm{0}, \mathbf{I}\right)$.

The reconstruction loss in Eq. \eqref{eq:recon_loss} corresponds to the negative log-likelihood (NLL) of the data sample given an assumed distribution and includes the decoder, which maps from latent space to a Gaussian distribution, from where we can sample the decoder output $\widehat{\mathbf{y}}$,
\begin{equation}
    \widehat{\mathbf{y}} \drawn \mathcal{N}({dec}_{\bm{\theta}}(\mathbf{z}_{x}), \mathbf{s}^2\identity).\label{eq:decoder_output}
\end{equation}
The standard deviation $\mathbf{s}$ is spatially shared with one value per output variable and learned alongside the autoencoder \cite{cipolla_multi-task_2018, rybkin_simple_2020, finn_deep_2023}.
From such a sample $\widehat{\mathbf{y}}$, we can recover the reconstructed data sample $\widehat{\mathbf{x}}$ by applying a deterministic (possibly non-invertible) transformation function $\widehat{\mathbf{x}} = f(\mathbf{\widehat{\mathbf{y}}})$.
Depending on the transformation function, we obtain different reconstruction losses:
\begin{itemize}
    \item{
        If we apply an identity function $f(\mathbf{\widehat{\mathbf{y}}}) = \mathbf{\widehat{\mathbf{y}}}$, the reconstruction loss is the NLL of a Gaussian distribution and includes a mean-squared error weighted by $\mathbf{s}$.
    }
    \item{
        If we clip the output into physical bounds, e.g. $f(\mathbf{\widehat{\mathbf{y}}}) = \mathrm{min}(\mathrm{max}(\mathbf{\widehat{\mathbf{y}}}, \mathbf{x}_{\mathsf{L}}), \mathbf{x}_{\mathsf{U}})$ by specifying a lower $\mathbf{x}_{\mathsf{L}}$ and/or upper bound
        $\mathbf{x}_{\mathsf{U}}$, the reconstruction loss is the NLL of a censored Gaussian distribution.
        The censored distribution utilises the decoder output, as described in Eq. \eqref{eq:decoder_output}, to regress the reconstruction and to determine whether it will be clipped into the bounds.
        Hence, this NLL combines a regression with a classification task and can be seen as NLL of a type-I Tobit model \cite{tobin_estimation_1958, amemiya_tobit_1984}.
    }
\end{itemize}
For a detailed treatment of the reconstruction loss, we refer to Appendix \ref{app:censored_dist}.

After training the autoencoder with Eq. \eqref{eq:autoencoder_loss}, we use the mean prediction of the encoder, $\mu_{\mathsf{enc}, \bm{\theta}}(\mathbf{x})$, as deterministic mapping from data space into latent space and train a diffusion model in this space.
The instantiated diffusion model \cite{sohl-dickstein_deep_2015, ho_denoising_2020} can be seen as denoiser $D_{\boldsymbol{\phi}}(\mathbf{z}_{\tau}, \tau)$ with its weights and biases $\boldsymbol{\phi}$ \cite{karras_elucidating_2022}, which takes a noised latent state $\mathbf{z}_{\tau}$ at a pseudo-time $\tau \in [0, 1]$ and should output the cleaned latent state $\mathbf{z}_{x}$.
During training, the noised latent state is produced by a \textit{variance-preserving} diffusion process \cite{kingma_variational_2021, song_maximum_2021}, under the assumption that the latent sample $\mathbf{z}_{x}$ has been normalised to mean $0$ and standard deviation $1$,
\begin{align}
    \mathbf{z}_{\tau} &= \alpha_{\tau} \mathbf{z}_{x} + \sigma_{\tau} \boldsymbol{\epsilon},  &\text{with}~\boldsymbol{\epsilon} \sim \mathcal{N}(\boldsymbol{0}, \mathbf{I}),\label{eq:diff_process}
\end{align}
where the signal $\alpha_{\tau}$ and noise $\sigma_{\tau}$ amplitudes are given in terms of logarithmic signal-to-noise ratio $\gamma(\tau) = \log\Big(\frac{\alpha_{\tau}^2}{\sigma_{\tau}^2}\Big)$.
The training noise scheduler specifies the dependency of the ratio on the pseudo-time and is adapted during the training process \cite{kingma_understanding_2023} as further described in Appendix \ref{app:noise_scheduler}.

Given a latent sample $\mathbf{z}_{x}$ and pseudo-time $\tau$, the loss function to train the diffusion model reads
\begin{align}
    \mathcal{L}(\mathbf{z}_{x}, \tau, \boldsymbol{\phi}) = w(\gamma)\Big(-\frac{d\gamma}{d\tau}\Big) e^{\gamma}\lVert \mathbf{z}_{x}- D_{\boldsymbol{\phi}}(\mathbf{z}_{\tau}, \tau)\rVert_{2}^{2},\label{eq:diff_loss}
\end{align}
with $w(\gamma) = \exp{(-\frac{\gamma}{2})}$ as external weighting function \cite{kingma_understanding_2023}.
The diffusion model is parameterised with a $v$-prediction \cite{salimans_progressive_2022} and implemented as UViT architecture \cite{hoogeboom_simple_2023} combining a U-Net \cite{ronneberger_u-net_2015} with a vision transformer \cite{dosovitskiy_image_2021}.
The minimum value of the noise scheduler is set to $\gamma_{\mathsf{min}}=-15$ and the maximum value to $\gamma_{\mathsf{max}}=15$.
We further describe the diffusion model in Appendix \ref{app:add_methods_diffusion} and its architecture in Appendix \ref{app:diff_architecture}.

To generate new data samples with the trained models, we draw a sample $\widehat{\mathbf{z}}_{1} \sim \mathcal{N}(\boldsymbol{0}, \mathbf{I})$, and integrate an ordinary differential equation \cite{song_maximum_2021} with our trained diffusion model $D_{\boldsymbol{\phi}}(\widehat{\mathbf{z}}_{\tau}, \tau)$ in $20$ integration steps. 
We use the second-order Heun integrator and sampling noise scheduler from \citet{karras_elucidating_2022} to obtain a latent sample $\widehat{\mathbf{z}}_{x}$.
We then apply the decoder to reconstruct a sample in data space $\widehat{\mathbf{x}}$.
If the decoder is trained with censoring, we clip the $\text{SIT} \in [0, \infty)$, $\text{SIC} \in [0, 1]$, and $\text{SID} \in [0, 1]$ to their physical bounds.

\section{Results}\label{sec:results}

In our experiments, we evaluate how LDMs perform compared to generative diffusion trained in data space and how the censoring of the decoder improves the physical consistency of the generated samples.
We train the models on simulations from the state-of-the-art Lagrangian sea-ice model neXtSIM \cite{rampal_nextsim_2016, olason_new_2022} which has been coupled to the ocean component of the NEMO framework \cite{madec_nemo_2008}.
The simulation data contains more than 20 simulation years \cite{boutin_arctic_2023} with a six-hourly output on a $\frac{1}{4}^{\circ}$ ($\simeq12 \,\text{km}$ in the Arctic) curvilinear grid.
Omitting the first five years as spin-up, we train the models on data from 2000--2014, validate on 2015, and estimate the here-presented test scores on 2016--2018.
All models are trained with ADAM \cite{kingma_adam_2017}, a batch size of $16$, and a learning rate of $2 \times 10^{-4}$ for $10^5$ iterations.
For further experimental details, we refer to Appendix \ref{app:exp_details}.

\begin{table}[ht]
\caption{
    Evaluation of the autoencoders for reconstructing the testing dataset with the normalised root-mean-squared error (RMSE), the structural similarity index measure (SSIM), and the accuracy in the sea-ice extent ($\text{ACC}_{\text{SIE}}$).
    $\beta$ is the regularisation factor.
}
\label{tab:autoencoder}
\begin{center}
\begin{small}
\begin{sc}
\begin{tabular}{lrrrrr}
\toprule
$\beta$ & RMSE $\downarrow$ & SSIM $\uparrow$ & $\text{ACC}_{\text{SIE}}$ $\uparrow$ \\
\midrule
$1$ & 0.123 & 0.878 & 0.964\\
$10^{-3}$ & \textbf{0.076}	& 0.939 & 0.986 \\
$10^{-3}$ (censored) & 0.078 & \textbf{0.941} & \textbf{0.991}\\
\bottomrule
\end{tabular}
\end{sc}
\end{small}
\end{center}
\end{table}

The autoencoders compress the input data from $5 \times 512 \times 512$ to $16 \times 64 \times 64$.
We ablate design choices and evaluate the autoencoder performance of the deterministic mapping as used for diffusion in terms of reconstruction quality with the mean squared error (MSE), the structural similarity index measure (SSIM), and the accuracy in the sea-ice extent ($\text{ACC}_{\text{SIE}}$).
The definition of the metrics and their calculations are explained in Appendix \ref{app:metrics}.
The results are shown in Table \ref{tab:autoencoder}.

Since its regularisation is much lower, the VAE trained with $\beta=10^{-3}$ reconstitutes into a better reconstruction than the VAE with $\beta=1$, resulting into lower errors and higher similarities.
Replacing the Gaussian log-likelihood in the reconstruction loss by a censored log-likelihood has a neutral impact on the RMSE and SSIM.
However, the VAE with censoring can better reconstruct the marginal ice zone, which leads to a higher accuracy in the sea-ice extent.
While higher accuracies therein might have only a small impact for generating new data, they might have a higher impact for other tasks where the output would be reused and were small errors can amplify, e.g., in surrogate modelling.

\begin{figure}[ht]
\begin{center}
\centerline{\includegraphics[width=0.45\textwidth]{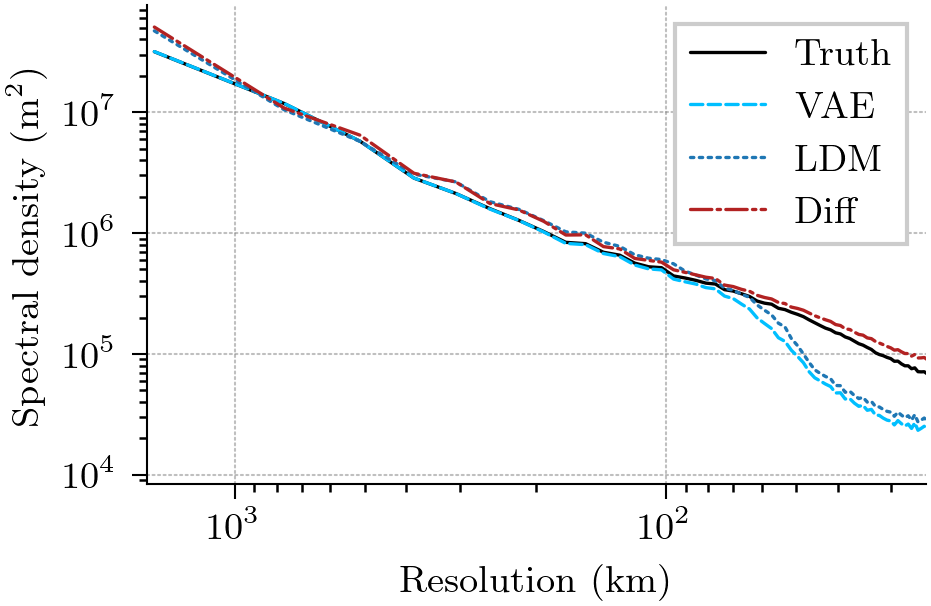}}
\caption{
    Estimated spectral density of the sea-ice thickness in the central Arctic for the testing dataset (black), an autoencoder ($\beta=10^{-3}$, no censoring, light blue), the LDM learned in the latent space of the autoencoder (dark blue), and the diffusion model directly learned in data space (red).
}
\label{fig:spectrum}
\end{center}
\vskip -0.3in
\end{figure}

On top of the pre-trained autoencoder, we train LDMs as described in Section \ref{sec:methods}.
We compare these LDMs to a diffusion model in data space, trained with Eq. \eqref{eq:diff_loss} by setting $\mathbf{z}_{x}=\mathbf{x}$.
As shown in Fig. \ref{fig:samples}, LDMs smooth the generated fields compared to diffusion in data space.
In the following, we establish how and why this smoothing happens by showing estimated spectra for the sea-ice thickness in Fig. \ref{fig:spectrum}.

LDMs lose small-scale information compared to the testing dataset, while diffusion models in data space retain information across all scales.
This loss of small-scale information results into a smoothing in the generated samples as shown in Fig. \ref{fig:samples}.
Since we find the same for the autoencoder, this smoothing is a result of the pre-training as variational autoencoder and the data compression in latent space.

The two diffusion models additionally experience a shift towards larger energies than the test dataset.
This shift might be caused by a discrepancy between training (2000--2014) and testing (2016--2018) or by the approximations during sampling.

\begin{table}[ht]
\caption{
    Evaluation of the statistics for samples generated by the diffusion models compared to the testing dataset with the same size.
    The Fréchet distance ($\mathrm{FAED}$) estimates how well the structures align while the RMSE in the sea-ice extent ($\text{RMSE}_{\text{SIE}}$) shows how well ice edges are represented, both are multiplied for visualisation by $10^{2}$.
    Diffusion corresponds to diffusion in data space, while LDM are the latent diffusion models.
    The validation dataset has only 1/3 of the testing dataset size.
}
\label{tab:diffusion}
\begin{center}
\begin{small}
\begin{sc}
\begin{tabular}{lrrr}
\toprule
Model & $\mathrm{FAED}$ $\downarrow$ & $\text{RMSE}_{\text{SIE}}$ $\downarrow$ & Speed ($\unit{s}$) $\downarrow$\\
\midrule
\textit{Validation}        & 2.38 & 7.02 & --\\
\midrule
Diffusion                  & 1.73 & 9.66 & 1.764\\
LDM                        & \textbf{1.65} & 11.56 & \textbf{0.069}\\
LDM (censored)             & 1.79 & \textbf{7.32} & 0.070\\
\bottomrule
\end{tabular}
\end{sc}
\end{small}
\end{center}
\end{table}

Inspired by the Fréchet inception distance \cite{heusel_gans_2017}, we evaluate the diffusion models in comparison to the testing dataset with the Fréchet distance in the latent space of an independently trained VAE ($\mathrm{FAED}$).
We additionally compare the probability that a grid point is covered by sea ice with the root-mean-squared error $\text{RMSE}_{\text{SIE}}$, with results shown in Table \ref{tab:diffusion}.
Whereas we evaluate on $4383$ generated samples only, the results are robust to the dataset size as demonstrated in Appendix \ref{app:results_dataset_size}.
Additionally, we measure the speed (latency) of the diffusion models in seconds for a single generation.

The diffusion models consistently outperform the validation dataset for the $\mathrm{FAED}$.
Although the validation $\mathrm{FAED}$ is limited by its dataset size, this nevertheless indicates that diffusion models effectively capture large-scale structures and correlations, as observed in the testing dataset.
The differences in the $\mathrm{FAED}$ between the diffusion models might be caused by the volatility during training.

Diffusion models are trained with a Gaussian diffusion process, and the data space model struggles to accurately represent the sea-ice extent, leading to a higher RMSE than for the validation dataset.
The smoothing in the LDM exacerbates the inaccuracies in representing sea-ice extent, resulting in even higher RMSE values.
However, as for the reconstruction, censoring improves the sea-ice extent representation, bringing the RMSE to the realm of the validation dataset.
In addition to an improved representation, the latent diffusion models are $25\times$ faster than the diffusion model in data space.

\section{Conclusions}\label{sec:conclusions}

In this manuscript, we make the first step towards using latent diffusion models (LDMs) for generating multivariate, Arctic-wide sea-ice data.
We compare LDMs with diffusion models in data space and find that LDMs lose small-scale information compared to the target fields, though their scores are similar to those of diffusion models in data space.
This loss of information is due to the latent mapping process and its pre-training as variational autoencoder.
To mitigate this, we could increase the number of channels in latent space (see Appendix \ref{app:results_ablation_ae}).
Another approach would be to pre-train the autoencoder with an additional adversarial \cite{rombach_high-resolution_2022} or spectral loss \cite{kochkov_neural_2023}.
These methods to try to mitigate the smoothing could complicate \citep[e.g.,][]{rombach_high-resolution_2022} and notoriously destabilise \citep[e.g.,][]{karras_progressive_2018} the training and tuning process.

Despite the loss of small-scale information, LDMs offer a significant advantage over diffusion models in data space: the ability to enforce physical bounds during data generation and their speed.
By replacing the Gaussian data log-likelihood by a censored log-likelihood, we explicitly incorporate the clipping of the decoder output into the training process.
The censoring has a neutral impact on the sample quality but leads to a more accurate representation of the sea-ice extent, indicating an increased physical consistency.
This increased consistency could benefit other tasks as surrogate modelling or downscaling.

While we have tested LDMs only for sea-ice data, we anticipate similar smoothing issues for other Earth system components.
Therefore, the smoothing issue remains a significant barrier to the broader application of the otherwise promising LDMs for large-scale Earth system modelling.

\section*{Acknowledgments}
This manuscript is a contribution to the DeepGeneSIS project as financed by INSU/CNRS in the PNTS program and the SASIP project, supported under Grant no. 353 by Schmidt Science, LLC -- a philanthropic initiative that seeks to improve societal outcomes through the development of emerging science and technologies.
TSF, CD, AF, MB additionally acknowledge financial support by INSU/CNRS for the project GenD$^2$M in the LEFE-MANU program.
This work was granted access to the HPC resources of IDRIS under the allocations 2023-AD011013069R2 made by GENCI.
We would like to thank Guillaume Boutin for providing access to the data and Laurent Bertino for helping to obtain the funding for the DeepGeneSIS project.
With their reviews, four anonymous referees have improved the manuscript.
CEREA is a member of the Institut Pierre-Simon Laplace (IPSL).

\section*{Impact Statement}

We present here work whose goal is to advance the field of Earth system modelling and machine learning.
While there are many potential societal consequences of our work, we will briefly discuss the impact on the energy consumption only.

Training neural networks consumes a lot of energy and in the future, the energy consumption of machine learning will rather increase than decrease, especially with the rise of generative models like diffusion models discussed here.
Finding methods to reduce the computational costs of those models can be key to also reduce their energy consumption.
While latent diffusion models might have the limitation of smoothing compared to diffusion models in data space, they offer an opportunity to decrease the costs and, thus, also the energy consumption.
This might be especially helpful if we further strive towards larger and larger deep learning architectures, in the end possibly for coupled Earth system models.

\bibliography{manuscript}

\begin{thebibliography}{67}
\providecommand{\natexlab}[1]{#1}
\providecommand{\url}[1]{\texttt{#1}}
\expandafter\ifx\csname urlstyle\endcsname\relax
  \providecommand{\doi}[1]{doi: #1}\else
  \providecommand{\doi}{doi: \begingroup \urlstyle{rm}\Url}\fi

\bibitem[Amemiya(1984)]{amemiya_tobit_1984}
Amemiya, T.
\newblock Tobit models: {A} survey.
\newblock \emph{Journal of Econometrics}, 24\penalty0 (1):\penalty0 3--61, January 1984.
\newblock ISSN 0304-4076.
\newblock \doi{10.1016/0304-4076(84)90074-5}.

\bibitem[Anderson(1982)]{anderson_reverse-time_1982}
Anderson, B. D.~O.
\newblock Reverse-time diffusion equation models.
\newblock \emph{Stochastic Processes and their Applications}, 12\penalty0 (3):\penalty0 313--326, May 1982.
\newblock ISSN 0304-4149.
\newblock \doi{10.1016/0304-4149(82)90051-5}.

\bibitem[Ba et~al.(2016)Ba, Kiros, and Hinton]{ba_layer_2016}
Ba, J.~L., Kiros, J.~R., and Hinton, G.~E.
\newblock Layer {Normalization}.
\newblock \emph{arXiv:1607.06450 [cs, stat]}, July 2016.
\newblock arXiv: 1607.06450 tex.ids= ba2016a.

\bibitem[Bar-Tal et~al.(2024)Bar-Tal, Chefer, Tov, Herrmann, Paiss, Zada, Ephrat, Hur, Liu, Raj, Li, Rubinstein, Michaeli, Wang, Sun, Dekel, and Mosseri]{bar-tal_lumiere_2024}
Bar-Tal, O., Chefer, H., Tov, O., Herrmann, C., Paiss, R., Zada, S., Ephrat, A., Hur, J., Liu, G., Raj, A., Li, Y., Rubinstein, M., Michaeli, T., Wang, O., Sun, D., Dekel, T., and Mosseri, I.
\newblock Lumiere: {A} {Space}-{Time} {Diffusion} {Model} for {Video} {Generation}, February 2024.
\newblock arXiv:2401.12945 [cs].

\bibitem[Bouchat et~al.(2022)Bouchat, Hutter, Chanut, Dupont, Dukhovskoy, Garric, Lee, Lemieux, Lique, Losch, Maslowski, Myers, Ólason, Rampal, Rasmussen, Talandier, Tremblay, and Wang]{bouchat_sea_2022}
Bouchat, A., Hutter, N., Chanut, J., Dupont, F., Dukhovskoy, D., Garric, G., Lee, Y.~J., Lemieux, J.-F., Lique, C., Losch, M., Maslowski, W., Myers, P.~G., Ólason, E., Rampal, P., Rasmussen, T., Talandier, C., Tremblay, B., and Wang, Q.
\newblock Sea {Ice} {Rheology} {Experiment} ({SIREx}): 1. {Scaling} and {Statistical} {Properties} of {Sea}-{Ice} {Deformation} {Fields}.
\newblock \emph{Journal of Geophysical Research: Oceans}, 127\penalty0 (4):\penalty0 e2021JC017667, 2022.
\newblock ISSN 2169-9291.
\newblock \doi{10.1029/2021JC017667}.
\newblock \_eprint: https://onlinelibrary.wiley.com/doi/pdf/10.1029/2021JC017667.

\bibitem[Boutin et~al.(2023)Boutin, Ólason, Rampal, Regan, Lique, Talandier, Brodeau, and Ricker]{boutin_arctic_2023}
Boutin, G., Ólason, E., Rampal, P., Regan, H., Lique, C., Talandier, C., Brodeau, L., and Ricker, R.
\newblock Arctic sea ice mass balance in a new coupled ice–ocean model using a brittle rheology framework.
\newblock \emph{The Cryosphere}, 17\penalty0 (2):\penalty0 617--638, February 2023.
\newblock ISSN 1994-0416.
\newblock \doi{10.5194/tc-17-617-2023}.
\newblock Publisher: Copernicus GmbH.

\bibitem[Brajard et~al.(2024)Brajard, Korosov, Davy, and Wang]{brajard_super-resolution_2024}
Brajard, J., Korosov, A., Davy, R., and Wang, Y.
\newblock Super-resolution of satellite observations of sea ice thickness using diffusion models and physical modeling.
\newblock Technical Report EGU24-3826, Copernicus Meetings, March 2024.
\newblock Conference Name: EGU24.

\bibitem[Cipolla et~al.(2018)Cipolla, Gal, and Kendall]{cipolla_multi-task_2018}
Cipolla, R., Gal, Y., and Kendall, A.
\newblock Multi-task {Learning} {Using} {Uncertainty} to {Weigh} {Losses} for {Scene} {Geometry} and {Semantics}.
\newblock In \emph{2018 {IEEE}/{CVF} {Conference} on {Computer} {Vision} and {Pattern} {Recognition}}, pp.\  7482--7491, Salt Lake City, UT, USA, June 2018. IEEE.
\newblock ISBN 978-1-5386-6420-9.
\newblock \doi{10.1109/CVPR.2018.00781}.

\bibitem[Dai et~al.(2023)Dai, Hou, Ma, Tsai, Wang, Wang, Zhang, Vandenhende, Wang, Dubey, Yu, Kadian, Radenovic, Mahajan, Li, Zhao, Petrovic, Singh, Motwani, Wen, Song, Sumbaly, Ramanathan, He, Vajda, and Parikh]{dai_emu_2023}
Dai, X., Hou, J., Ma, C.-Y., Tsai, S., Wang, J., Wang, R., Zhang, P., Vandenhende, S., Wang, X., Dubey, A., Yu, M., Kadian, A., Radenovic, F., Mahajan, D., Li, K., Zhao, Y., Petrovic, V., Singh, M.~K., Motwani, S., Wen, Y., Song, Y., Sumbaly, R., Ramanathan, V., He, Z., Vajda, P., and Parikh, D.
\newblock Emu: {Enhancing} {Image} {Generation} {Models} {Using} {Photogenic} {Needles} in a {Haystack}, September 2023.
\newblock arXiv:2309.15807 [cs].

\bibitem[Dansereau et~al.(2016)Dansereau, Weiss, Saramito, and Lattes]{dansereau_maxwell_2016}
Dansereau, V., Weiss, J., Saramito, P., and Lattes, P.
\newblock A {Maxwell} elasto-brittle rheology for sea ice modelling.
\newblock \emph{The Cryosphere}, 10\penalty0 (3):\penalty0 1339--1359, July 2016.
\newblock ISSN 1994-0416.
\newblock \doi{10.5194/tc-10-1339-2016}.
\newblock publisher: Copernicus GmbH.

\bibitem[Dhariwal \& Nichol(2021)Dhariwal and Nichol]{dhariwal_diffusion_2021}
Dhariwal, P. and Nichol, A.
\newblock Diffusion {Models} {Beat} {GANs} on {Image} {Synthesis}.
\newblock In \emph{Advances in {Neural} {Information} {Processing} {Systems}}, volume~34, pp.\  8780--8794. Curran Associates, Inc., 2021.

\bibitem[Dosovitskiy et~al.(2021)Dosovitskiy, Beyer, Kolesnikov, Weissenborn, Zhai, Unterthiner, Dehghani, Minderer, Heigold, Gelly, Uszkoreit, and Houlsby]{dosovitskiy_image_2021}
Dosovitskiy, A., Beyer, L., Kolesnikov, A., Weissenborn, D., Zhai, X., Unterthiner, T., Dehghani, M., Minderer, M., Heigold, G., Gelly, S., Uszkoreit, J., and Houlsby, N.
\newblock An {Image} is {Worth} 16x16 {Words}: {Transformers} for {Image} {Recognition} at {Scale}, June 2021.
\newblock arXiv:2010.11929 [cs].

\bibitem[Durand et~al.(2024)Durand, Finn, Farchi, Bocquet, Boutin, and Ólason]{durand_data-driven_2024}
Durand, C., Finn, T.~S., Farchi, A., Bocquet, M., Boutin, G., and Ólason, E.
\newblock Data-driven surrogate modeling of high-resolution sea-ice thickness in the {Arctic}.
\newblock \emph{The Cryosphere}, 18\penalty0 (4):\penalty0 1791--1815, April 2024.
\newblock ISSN 1994-0416.
\newblock \doi{10.5194/tc-18-1791-2024}.
\newblock Publisher: Copernicus GmbH.

\bibitem[Falcon et~al.(2020)Falcon, Borovec, Wälchli, Eggert, Schock, Jordan, Skafte, Ir1dXD, Bereznyuk, Harris, Murrell, Yu, Præsius, Addair, Zhong, Lipin, Uchida, Bapat, Schröter, Dayma, Karnachev, Kulkarni, Komatsu, Martin.B, SCHIRATTI, Mary, Byrne, Eyzaguirre, cinjon, and Bakhtin]{falcon_pytorchlightning_2020}
Falcon, W., Borovec, J., Wälchli, A., Eggert, N., Schock, J., Jordan, J., Skafte, N., Ir1dXD, Bereznyuk, V., Harris, E., Murrell, T., Yu, P., Præsius, S., Addair, T., Zhong, J., Lipin, D., Uchida, S., Bapat, S., Schröter, H., Dayma, B., Karnachev, A., Kulkarni, A., Komatsu, S., Martin.B, SCHIRATTI, J.-B., Mary, H., Byrne, D., Eyzaguirre, C., cinjon, and Bakhtin, A.
\newblock {PyTorchLightning}: 0.7.6 release, May 2020.

\bibitem[Finn et~al.(2023)Finn, Durand, Farchi, Bocquet, Chen, Carrassi, and Dansereau]{finn_deep_2023}
Finn, T.~S., Durand, C., Farchi, A., Bocquet, M., Chen, Y., Carrassi, A., and Dansereau, V.
\newblock Deep learning subgrid-scale parametrisations for short-term forecasting of sea-ice dynamics with a {Maxwell} elasto-brittle rheology.
\newblock \emph{The Cryosphere}, 17\penalty0 (7):\penalty0 2965--2991, July 2023.
\newblock ISSN 1994-0416.
\newblock \doi{10.5194/tc-17-2965-2023}.
\newblock Publisher: Copernicus GmbH.

\bibitem[Finn et~al.(2024)Finn, Durand, Farchi, Bocquet, Rampal, and Carrassi]{finn_generative_2024}
Finn, T.~S., Durand, C., Farchi, A., Bocquet, M., Rampal, P., and Carrassi, A.
\newblock Generative diffusion for regional surrogate models from sea-ice simulations.
\newblock April 2024.
\newblock \doi{10.22541/au.171386536.64344222/v1}.

\bibitem[Fishman et~al.(2023)Fishman, Klarner, Bortoli, Mathieu, and Hutchinson]{fishman_diffusion_2023}
Fishman, N., Klarner, L., Bortoli, V.~D., Mathieu, E., and Hutchinson, M.~J.
\newblock Diffusion models for constrained domains.
\newblock \emph{Transactions on Machine Learning Research}, 2023.
\newblock ISSN 2835-8856.

\bibitem[Girard et~al.(2011)Girard, Bouillon, Weiss, Amitrano, Fichefet, and Legat]{girard_new_2011}
Girard, L., Bouillon, S., Weiss, J., Amitrano, D., Fichefet, T., and Legat, V.
\newblock A new modeling framework for sea-ice mechanics based on elasto-brittle rheology.
\newblock \emph{Annals of Glaciology}, 52\penalty0 (57):\penalty0 123--132, January 2011.
\newblock ISSN 0260-3055, 1727-5644.
\newblock \doi{10.3189/172756411795931499}.

\bibitem[Goessling et~al.(2016)Goessling, Tietsche, Day, Hawkins, and Jung]{goessling_predictability_2016}
Goessling, H.~F., Tietsche, S., Day, J.~J., Hawkins, E., and Jung, T.
\newblock Predictability of the {Arctic} sea ice edge.
\newblock \emph{Geophysical Research Letters}, 43\penalty0 (4):\penalty0 1642--1650, 2016.
\newblock ISSN 1944-8007.
\newblock \doi{10.1002/2015GL067232}.
\newblock \_eprint: https://onlinelibrary.wiley.com/doi/pdf/10.1002/2015GL067232.

\bibitem[Hendrycks \& Gimpel(2016)Hendrycks and Gimpel]{hendrycks_gaussian_2016}
Hendrycks, D. and Gimpel, K.
\newblock Gaussian error linear units (gelus), 2016.

\bibitem[Hersbach et~al.(2020)Hersbach, Bell, Berrisford, Hirahara, Horányi, Muñoz-Sabater, Nicolas, Peubey, Radu, Schepers, Simmons, Soci, Abdalla, Abellan, Balsamo, Bechtold, Biavati, Bidlot, Bonavita, Chiara, Dahlgren, Dee, Diamantakis, Dragani, Flemming, Forbes, Fuentes, Geer, Haimberger, Healy, Hogan, Hólm, Janisková, Keeley, Laloyaux, Lopez, Lupu, Radnoti, Rosnay, Rozum, Vamborg, Villaume, and Thépaut]{hersbach_era5_2020}
Hersbach, H., Bell, B., Berrisford, P., Hirahara, S., Horányi, A., Muñoz-Sabater, J., Nicolas, J., Peubey, C., Radu, R., Schepers, D., Simmons, A., Soci, C., Abdalla, S., Abellan, X., Balsamo, G., Bechtold, P., Biavati, G., Bidlot, J., Bonavita, M., Chiara, G.~D., Dahlgren, P., Dee, D., Diamantakis, M., Dragani, R., Flemming, J., Forbes, R., Fuentes, M., Geer, A., Haimberger, L., Healy, S., Hogan, R.~J., Hólm, E., Janisková, M., Keeley, S., Laloyaux, P., Lopez, P., Lupu, C., Radnoti, G., Rosnay, P.~d., Rozum, I., Vamborg, F., Villaume, S., and Thépaut, J.-N.
\newblock The {ERA5} global reanalysis.
\newblock \emph{Quarterly Journal of the Royal Meteorological Society}, 146\penalty0 (730):\penalty0 1999--2049, 2020.
\newblock ISSN 1477-870X.
\newblock \doi{https://doi.org/10.1002/qj.3803}.
\newblock \_eprint: https://onlinelibrary.wiley.com/doi/pdf/10.1002/qj.3803.

\bibitem[Heusel et~al.(2017)Heusel, Ramsauer, Unterthiner, Nessler, and Hochreiter]{heusel_gans_2017}
Heusel, M., Ramsauer, H., Unterthiner, T., Nessler, B., and Hochreiter, S.
\newblock {GANs} {Trained} by a {Two} {Time}-{Scale} {Update} {Rule} {Converge} to a {Local} {Nash} {Equilibrium}.
\newblock \emph{arXiv:1706.08500 [cs, stat]}, June 2017.
\newblock arXiv: 1706.08500.

\bibitem[Higgins et~al.(2016)Higgins, Matthey, Pal, Burgess, Glorot, Botvinick, Mohamed, and Lerchner]{higgins_beta-vae:_2016}
Higgins, I., Matthey, L., Pal, A., Burgess, C., Glorot, X., Botvinick, M., Mohamed, S., and Lerchner, A.
\newblock beta-{VAE}: {Learning} {Basic} {Visual} {Concepts} with a {Constrained} {Variational} {Framework}.
\newblock November 2016.

\bibitem[Ho et~al.(2020)Ho, Jain, and Abbeel]{ho_denoising_2020}
Ho, J., Jain, A., and Abbeel, P.
\newblock Denoising {Diffusion} {Probabilistic} {Models}, December 2020.
\newblock tex.ids= ho2020, ho2020a, ho2020b arXiv: 2006.11239 [cs, stat] number: arXiv:2006.11239.

\bibitem[Hoogeboom et~al.(2023)Hoogeboom, Heek, and Salimans]{hoogeboom_simple_2023}
Hoogeboom, E., Heek, J., and Salimans, T.
\newblock simple diffusion: {End}-to-end diffusion for high resolution images, January 2023.
\newblock arXiv:2301.11093 [cs, stat].

\bibitem[Huang et~al.(2023)Huang, Park, Wang, Denk, Ly, Chen, Zhang, Zhang, Yu, Frank, Engel, Le, Chan, Chen, and Han]{huang_noise2music_2023}
Huang, Q., Park, D.~S., Wang, T., Denk, T.~I., Ly, A., Chen, N., Zhang, Z., Zhang, Z., Yu, J., Frank, C., Engel, J., Le, Q.~V., Chan, W., Chen, Z., and Han, W.
\newblock {Noise2Music}: {Text}-conditioned {Music} {Generation} with {Diffusion} {Models}, March 2023.
\newblock arXiv:2302.03917 [cs, eess].

\bibitem[Kadow et~al.(2020)Kadow, Hall, and Ulbrich]{kadow_artificial_2020}
Kadow, C., Hall, D.~M., and Ulbrich, U.
\newblock Artificial intelligence reconstructs missing climate information.
\newblock \emph{Nature Geoscience}, 13\penalty0 (6):\penalty0 408--413, June 2020.
\newblock ISSN 1752-0908.
\newblock \doi{10.1038/s41561-020-0582-5}.
\newblock Number: 6 Publisher: Nature Publishing Group.

\bibitem[Karras et~al.(2018)Karras, Aila, Laine, and Lehtinen]{karras_progressive_2018}
Karras, T., Aila, T., Laine, S., and Lehtinen, J.
\newblock Progressive {Growing} of {GANs} for {Improved} {Quality}, {Stability}, and {Variation}.
\newblock February 2018.

\bibitem[Karras et~al.(2022)Karras, Aittala, Aila, and Laine]{karras_elucidating_2022}
Karras, T., Aittala, M., Aila, T., and Laine, S.
\newblock Elucidating the {Design} {Space} of {Diffusion}-{Based} {Generative} {Models}, October 2022.
\newblock arXiv:2206.00364 [cs, stat].

\bibitem[Kingma et~al.(2021)Kingma, Salimans, Poole, and Ho]{kingma_variational_2021}
Kingma, D., Salimans, T., Poole, B., and Ho, J.
\newblock Variational {Diffusion} {Models}.
\newblock In \emph{Advances in {Neural} {Information} {Processing} {Systems}}, volume~34, pp.\  21696--21707. Curran Associates, Inc., 2021.
\newblock tex.ids= kingma\_variational\_2021-1.

\bibitem[Kingma \& Ba(2017)Kingma and Ba]{kingma_adam_2017}
Kingma, D.~P. and Ba, J.
\newblock Adam: {A} {Method} for {Stochastic} {Optimization}.
\newblock \emph{arXiv:1412.6980 [cs]}, January 2017.
\newblock arXiv: 1412.6980.

\bibitem[Kingma \& Gao(2023)Kingma and Gao]{kingma_understanding_2023}
Kingma, D.~P. and Gao, R.
\newblock Understanding {Diffusion} {Objectives} as the {ELBO} with {Simple} {Data} {Augmentation}, September 2023.
\newblock arXiv:2303.00848 [cs, stat].

\bibitem[Kingma \& Welling(2013)Kingma and Welling]{kingma_auto-encoding_2013}
Kingma, D.~P. and Welling, M.
\newblock Auto-{Encoding} {Variational} {Bayes}.
\newblock \emph{arXiv:1312.6114 [cs, stat]}, December 2013.
\newblock arXiv: 1312.6114.

\bibitem[Kochkov et~al.(2023)Kochkov, Yuval, Langmore, Norgaard, Smith, Mooers, Lottes, Rasp, Düben, Klöwer, Hatfield, Battaglia, Sanchez-Gonzalez, Willson, Brenner, and Hoyer]{kochkov_neural_2023}
Kochkov, D., Yuval, J., Langmore, I., Norgaard, P., Smith, J., Mooers, G., Lottes, J., Rasp, S., Düben, P., Klöwer, M., Hatfield, S., Battaglia, P., Sanchez-Gonzalez, A., Willson, M., Brenner, M.~P., and Hoyer, S.
\newblock Neural {General} {Circulation} {Models}, November 2023.
\newblock arXiv:2311.07222 [physics].

\bibitem[Leinonen et~al.(2023)Leinonen, Hamann, Nerini, Germann, and Franch]{leinonen_latent_2023}
Leinonen, J., Hamann, U., Nerini, D., Germann, U., and Franch, G.
\newblock Latent diffusion models for generative precipitation nowcasting with accurate uncertainty quantification, April 2023.
\newblock arXiv:2304.12891 [physics].

\bibitem[Li et~al.(2023)Li, Carver, Lopez-Gomez, Sha, and Anderson]{li_seeds_2023}
Li, L., Carver, R., Lopez-Gomez, I., Sha, F., and Anderson, J.
\newblock {SEEDS}: {Emulation} of {Weather} {Forecast} {Ensembles} with {Diffusion} {Models}, October 2023.
\newblock arXiv:2306.14066 [physics].

\bibitem[Liu et~al.(2018)Liu, Reda, Shih, Wang, Tao, and Catanzaro]{liu_image_2018}
Liu, G., Reda, F.~A., Shih, K.~J., Wang, T.-C., Tao, A., and Catanzaro, B.
\newblock Image {Inpainting} for {Irregular} {Holes} {Using} {Partial} {Convolutions}.
\newblock pp.\  85--100, 2018.

\bibitem[Liu et~al.(2022)Liu, Mao, Wu, Feichtenhofer, Darrell, and Xie]{liu_convnet_2022}
Liu, Z., Mao, H., Wu, C.-Y., Feichtenhofer, C., Darrell, T., and Xie, S.
\newblock A {ConvNet} for the 2020s, March 2022.

\bibitem[Lou \& Ermon(2023)Lou and Ermon]{lou_reflected_2023}
Lou, A. and Ermon, S.
\newblock Reflected {Diffusion} {Models}.
\newblock In \emph{Proceedings of the 40th {International} {Conference} on {Machine} {Learning}}, pp.\  22675--22701. PMLR, July 2023.
\newblock ISSN: 2640-3498.

\bibitem[Madec(2008)]{madec_nemo_2008}
Madec, G.
\newblock {NEMO} ocean engine.
\newblock Project report 1288-1619, Institut Pierre-Simon Laplace (IPSL), 2008.
\newblock Series: 27.

\bibitem[Mardani et~al.(2023)Mardani, Brenowitz, Cohen, Pathak, Chen, Liu, Vahdat, Kashinath, Kautz, and Pritchard]{mardani_generative_2023}
Mardani, M., Brenowitz, N., Cohen, Y., Pathak, J., Chen, C.-Y., Liu, C.-C., Vahdat, A., Kashinath, K., Kautz, J., and Pritchard, M.
\newblock Generative {Residual} {Diffusion} {Modeling} for {Km}-scale {Atmospheric} {Downscaling}, September 2023.
\newblock arXiv:2309.15214 [physics].

\bibitem[Odena et~al.(2016)Odena, Dumoulin, and Olah]{odena_deconvolution_2016}
Odena, A., Dumoulin, V., and Olah, C.
\newblock Deconvolution and {Checkerboard} {Artifacts}.
\newblock \emph{Distill}, 1\penalty0 (10):\penalty0 e3, October 2016.
\newblock ISSN 2476-0757.
\newblock \doi{10.23915/distill.00003}.

\bibitem[Paszke et~al.(2019)Paszke, Gross, Massa, Lerer, Bradbury, Chanan, Killeen, Lin, Gimelshein, Antiga, Desmaison, Kopf, Yang, DeVito, Raison, Tejani, Chilamkurthy, Steiner, Fang, Bai, and Chintala]{paszke_pytorch_2019}
Paszke, A., Gross, S., Massa, F., Lerer, A., Bradbury, J., Chanan, G., Killeen, T., Lin, Z., Gimelshein, N., Antiga, L., Desmaison, A., Kopf, A., Yang, E., DeVito, Z., Raison, M., Tejani, A., Chilamkurthy, S., Steiner, B., Fang, L., Bai, J., and Chintala, S.
\newblock {PyTorch}: {An} {Imperative} {Style}, {High}-{Performance} {Deep} {Learning} {Library}.
\newblock In Wallach, H., Larochelle, H., Beygelzimer, A., Alché-Buc, F., Fox, E., and Garnett, R. (eds.), \emph{Advances in {Neural} {Information} {Processing} {Systems} 32}, pp.\  8024--8035. Curran Associates, Inc., 2019.

\bibitem[Perez et~al.(2017)Perez, Strub, de~Vries, Dumoulin, and Courville]{perez_film_2017}
Perez, E., Strub, F., de~Vries, H., Dumoulin, V., and Courville, A.
\newblock {FiLM}: {Visual} {Reasoning} with a {General} {Conditioning} {Layer}, December 2017.
\newblock arXiv:1709.07871 [cs, stat].

\bibitem[Price et~al.(2024)Price, Sanchez-Gonzalez, Alet, Andersson, El-Kadi, Masters, Ewalds, Stott, Mohamed, Battaglia, Lam, and Willson]{price_gencast_2024}
Price, I., Sanchez-Gonzalez, A., Alet, F., Andersson, T.~R., El-Kadi, A., Masters, D., Ewalds, T., Stott, J., Mohamed, S., Battaglia, P., Lam, R., and Willson, M.
\newblock {GenCast}: {Diffusion}-based ensemble forecasting for medium-range weather, May 2024.
\newblock arXiv:2312.15796 [physics] version: 2.

\bibitem[Rampal et~al.(2016)Rampal, Bouillon, Ólason, and Morlighem]{rampal_nextsim_2016}
Rampal, P., Bouillon, S., Ólason, E., and Morlighem, M.
\newblock {neXtSIM}: a new {Lagrangian} sea ice model.
\newblock \emph{The Cryosphere}, 10\penalty0 (3):\penalty0 1055--1073, May 2016.
\newblock ISSN 1994-0416.
\newblock \doi{10.5194/tc-10-1055-2016}.
\newblock publisher: Copernicus GmbH.

\bibitem[Rampal et~al.(2019)Rampal, Dansereau, Olason, Bouillon, Williams, Korosov, and Samaké]{rampal_multi-fractal_2019}
Rampal, P., Dansereau, V., Olason, E., Bouillon, S., Williams, T., Korosov, A., and Samaké, A.
\newblock On the multi-fractal scaling properties of sea ice deformation.
\newblock \emph{The Cryosphere}, 13\penalty0 (9):\penalty0 2457--2474, September 2019.
\newblock ISSN 1994-0416.
\newblock \doi{10.5194/tc-13-2457-2019}.
\newblock Publisher: Copernicus GmbH.

\bibitem[Rombach et~al.(2022)Rombach, Blattmann, Lorenz, Esser, and Ommer]{rombach_high-resolution_2022}
Rombach, R., Blattmann, A., Lorenz, D., Esser, P., and Ommer, B.
\newblock High-{Resolution} {Image} {Synthesis} {With} {Latent} {Diffusion} {Models}.
\newblock pp.\  10684--10695, 2022.

\bibitem[Ronneberger et~al.(2015)Ronneberger, Fischer, and Brox]{ronneberger_u-net_2015}
Ronneberger, O., Fischer, P., and Brox, T.
\newblock U-{Net}: {Convolutional} {Networks} for {Biomedical} {Image} {Segmentation}.
\newblock \emph{arXiv:1505.04597 [cs]}, May 2015.
\newblock arXiv: 1505.04597.

\bibitem[Rybkin et~al.(2020)Rybkin, Daniilidis, and Levine]{rybkin_simple_2020}
Rybkin, O., Daniilidis, K., and Levine, S.
\newblock Simple and {Effective} {VAE} {Training} with {Calibrated} {Decoders}.
\newblock \emph{arXiv:2006.13202 [cs, eess, stat]}, June 2020.
\newblock arXiv: 2006.13202.

\bibitem[Saharia et~al.(2022{\natexlab{a}})Saharia, Chan, Saxena, Li, Whang, Denton, Ghasemipour, Gontijo~Lopes, Karagol~Ayan, Salimans, Ho, Fleet, and Norouzi]{saharia_photorealistic_2022}
Saharia, C., Chan, W., Saxena, S., Li, L., Whang, J., Denton, E.~L., Ghasemipour, K., Gontijo~Lopes, R., Karagol~Ayan, B., Salimans, T., Ho, J., Fleet, D.~J., and Norouzi, M.
\newblock Photorealistic {Text}-to-{Image} {Diffusion} {Models} with {Deep} {Language} {Understanding}.
\newblock \emph{Advances in Neural Information Processing Systems}, 35:\penalty0 36479--36494, December 2022{\natexlab{a}}.

\bibitem[Saharia et~al.(2022{\natexlab{b}})Saharia, Ho, Chan, Salimans, Fleet, and Norouzi]{saharia_image_2022}
Saharia, C., Ho, J., Chan, W., Salimans, T., Fleet, D.~J., and Norouzi, M.
\newblock Image {Super}-{Resolution} {Via} {Iterative} {Refinement}.
\newblock \emph{IEEE Transactions on Pattern Analysis and Machine Intelligence}, pp.\  1--14, 2022{\natexlab{b}}.
\newblock ISSN 1939-3539.
\newblock \doi{10.1109/TPAMI.2022.3204461}.
\newblock Conference Name: IEEE Transactions on Pattern Analysis and Machine Intelligence.

\bibitem[Salimans \& Ho(2022)Salimans and Ho]{salimans_progressive_2022}
Salimans, T. and Ho, J.
\newblock Progressive {Distillation} for {Fast} {Sampling} of {Diffusion} {Models}, June 2022.
\newblock arXiv:2202.00512 [cs, stat].

\bibitem[Shmakov et~al.(2023)Shmakov, Greif, Fenton, Ghosh, Baldi, and Whiteson]{shmakov_end--end_2023}
Shmakov, A., Greif, K., Fenton, M., Ghosh, A., Baldi, P., and Whiteson, D.
\newblock End-{To}-{End} {Latent} {Variational} {Diffusion} {Models} for {Inverse} {Problems} in {High} {Energy} {Physics}, May 2023.
\newblock arXiv:2305.10399 [hep-ex].

\bibitem[Sinha et~al.(2021)Sinha, Song, Meng, and Ermon]{sinha_d2c_2021}
Sinha, A., Song, J., Meng, C., and Ermon, S.
\newblock {D2C}: {Diffusion}-{Decoding} {Models} for {Few}-{Shot} {Conditional} {Generation}.
\newblock In \emph{Advances in {Neural} {Information} {Processing} {Systems}}, volume~34, pp.\  12533--12548. Curran Associates, Inc., 2021.

\bibitem[Sohl-Dickstein et~al.(2015)Sohl-Dickstein, Weiss, Maheswaranathan, and Ganguli]{sohl-dickstein_deep_2015}
Sohl-Dickstein, J., Weiss, E.~A., Maheswaranathan, N., and Ganguli, S.
\newblock Deep {Unsupervised} {Learning} using {Nonequilibrium} {Thermodynamics}.
\newblock \emph{arXiv:1503.03585 [cond-mat, q-bio, stat]}, November 2015.
\newblock arXiv: 1503.03585.

\bibitem[Song \& Ermon(2020)Song and Ermon]{song_improved_2020}
Song, Y. and Ermon, S.
\newblock Improved {Techniques} for {Training} {Score}-{Based} {Generative} {Models}.
\newblock In \emph{Advances in {Neural} {Information} {Processing} {Systems}}, volume~33, pp.\  12438--12448. Curran Associates, Inc., 2020.

\bibitem[Song et~al.(2021)Song, Durkan, Murray, and Ermon]{song_maximum_2021}
Song, Y., Durkan, C., Murray, I., and Ermon, S.
\newblock Maximum {Likelihood} {Training} of {Score}-{Based} {Diffusion} {Models}, October 2021.
\newblock arXiv:2101.09258 [cs, stat].

\bibitem[Talandier \& Lique(2021)Talandier and Lique]{talandier_creg025l75-nemo_r360_2021}
Talandier, C. and Lique, C.
\newblock {CREG025}.{L75}-{NEMO}\_r3.6.0, December 2021.

\bibitem[Tobin(1958)]{tobin_estimation_1958}
Tobin, J.
\newblock Estimation of {Relationships} for {Limited} {Dependent} {Variables}.
\newblock \emph{Econometrica}, 26\penalty0 (1):\penalty0 24--36, 1958.
\newblock ISSN 0012-9682.
\newblock \doi{10.2307/1907382}.
\newblock Publisher: [Wiley, Econometric Society].

\bibitem[Vahdat et~al.(2021)Vahdat, Kreis, and Kautz]{vahdat_score-based_2021}
Vahdat, A., Kreis, K., and Kautz, J.
\newblock Score-based {Generative} {Modeling} in {Latent} {Space}, December 2021.
\newblock arXiv:2106.05931 [cs, stat].

\bibitem[Vaswani et~al.(2017)Vaswani, Shazeer, Parmar, Uszkoreit, Jones, Gomez, Kaiser, and Polosukhin]{vaswani_attention_2017}
Vaswani, A., Shazeer, N., Parmar, N., Uszkoreit, J., Jones, L., Gomez, A.~N., Kaiser, L., and Polosukhin, I.
\newblock Attention {Is} {All} {You} {Need}.
\newblock \emph{arXiv:1706.03762 [cs]}, December 2017.
\newblock tex.ids: vaswani\_attention\_2017-1 arXiv: 1706.03762.

\bibitem[Wan et~al.(2023)Wan, Baptista, Chen, Anderson, Boral, Sha, and Zepeda-Núñez]{wan_debias_2023}
Wan, Z.~Y., Baptista, R., Chen, Y.-f., Anderson, J., Boral, A., Sha, F., and Zepeda-Núñez, L.
\newblock Debias {Coarsely}, {Sample} {Conditionally}: {Statistical} {Downscaling} through {Optimal} {Transport} and {Probabilistic} {Diffusion} {Models}, October 2023.
\newblock arXiv:2305.15618 [physics].

\bibitem[Xiong et~al.(2020)Xiong, Yang, He, Zheng, Zheng, Xing, Zhang, Lan, Wang, and Liu]{xiong_layer_2020}
Xiong, R., Yang, Y., He, D., Zheng, K., Zheng, S., Xing, C., Zhang, H., Lan, Y., Wang, L., and Liu, T.
\newblock On {Layer} {Normalization} in the {Transformer} {Architecture}.
\newblock In \emph{International {Conference} on {Machine} {Learning}}, pp.\  10524--10533. PMLR, November 2020.
\newblock ISSN: 2640-3498.

\bibitem[Yadan(2019)]{yadan_hydra_2019}
Yadan, O.
\newblock Hydra - {A} framework for elegantly configuring complex applications, 2019.
\newblock tex.howpublished: Github.

\bibitem[Ólason et~al.(2021)Ólason, Rampal, and Dansereau]{olason_statistical_2021}
Ólason, E., Rampal, P., and Dansereau, V.
\newblock On the statistical properties of sea-ice lead fraction and heat fluxes in the {Arctic}.
\newblock \emph{The Cryosphere}, 15\penalty0 (2):\penalty0 1053--1064, February 2021.
\newblock ISSN 1994-0416.
\newblock \doi{10.5194/tc-15-1053-2021}.
\newblock Publisher: Copernicus GmbH.

\bibitem[Ólason et~al.(2022)Ólason, Boutin, Korosov, Rampal, Williams, Kimmritz, Dansereau, and Samaké]{olason_new_2022}
Ólason, E., Boutin, G., Korosov, A., Rampal, P., Williams, T., Kimmritz, M., Dansereau, V., and Samaké, A.
\newblock A {New} {Brittle} {Rheology} and {Numerical} {Framework} for {Large}-{Scale} {Sea}-{Ice} {Models}.
\newblock \emph{Journal of Advances in Modeling Earth Systems}, 14\penalty0 (8):\penalty0 e2021MS002685, 2022.
\newblock ISSN 1942-2466.
\newblock \doi{10.1029/2021MS002685}.
\newblock \_eprint: https://onlinelibrary.wiley.com/doi/pdf/10.1029/2021MS002685.

\end{thebibliography}
\bibliographystyle{icml2024}

\newpage
\appendix
\onecolumn

\section{Additional methods}\label{app:add_methods}

In the main manuscript, we shortly explained the methods around our main message: we can train latent diffusion models for geophysics end-to-end and incorporate physical bounds into the latent diffusion model.
In the following, we explain the methods more in detail.
We elucidate on our variational diffusion formulation in Appendix \ref{app:add_methods_diffusion}.
We introduce censored Gaussian distributions for the data log-likelihood in Appendix \ref{app:censored_dist}, and we discuss our noise scheduler in Appendix \ref{app:noise_scheduler}.

\subsection{Latent diffusion formulation}\label{app:add_methods_diffusion}

In our formulation of generative diffusion, we rely on variational diffusion models as introduced in \citet{kingma_variational_2021}.
These models allow us to naturally make use of encoder and decoder structures to reduce the dimensionality of the system.

Given data drawn from a dataset $\mathbf{x} \sim \mathcal{D}$, the encoder produces the corresponding latent state $\mathbf{z}_{x} = \mathrm{enc}(\mathbf{x})$.
This latent state is noised by a \textit{variance-preserving} diffusion process,
\begin{align}
     \mathbf{z}_{\tau} &= \alpha_{\tau} \mathbf{z}_{x} + \sigma_{\tau} \boldsymbol{\epsilon},  &\text{with}~\boldsymbol{\epsilon} \sim \mathcal{N}(\boldsymbol{0}, \mathbf{I}),\tag{\ref{eq:diff_process}}
\end{align}
where the pseudo-time $\tau \in [0, 1]$ determines the temporal position of the process.
The signal amplitude $\alpha_{\tau}$ specifies how much signal is contained in the noised sample, and the noise amplitude $\sigma_{\tau}=\sqrt{1-\alpha_{\tau}^2}$ how much signal has been replaced by noise.
The dependency of the (noise) scheduling of $\alpha_{\tau}$ and $\sigma_{\tau}$ on the pseudo-time is formulated as logarithmic signal-to-noise ratio $\gamma(\tau) = \log(\frac{\alpha_{\tau}^2}{\sigma_{\tau}^2})$ such that we can recover $\alpha_{\tau}^2 = \frac{1}{1+\exp(-\gamma(\tau))}$ and $\sigma_{\tau}^2 = \frac{1}{1+\exp(\gamma(\tau))}$.

As in the variational autoencoder framework \cite{kingma_auto-encoding_2013}, the data log-likelihood $\log(p(\mathbf{x}))$ is lower bounded by the evidence lower bound (ELBO).
In equivalence, we can write an upper bound on the negative data log-likelihood \cite{kingma_variational_2021},
\begin{equation}
    -\log(p(\mathbf{x})) \leq -\expect_{\mathbf{z}_{0} \drawn q(\mathbf{z}_{0} \mid \mathbf{x})} \Big[\log p(\mathbf{x} \mid \mathbf{z}_{0}) \Big] + \kl[\bigg]{q(\mathbf{z}_{1} \mid \mathbf{x})}{p(\mathbf{z}_{1})} + \expect_{\mathbf{z}_{x} \drawn q(\mathbf{z}_{x} \mid \mathbf{x}), \tau \drawn [0, 1]}\mathcal{L}(\phi, \mathbf{z}_{x}, \tau).\label{eq:elbo}
\end{equation}
The first term is the negative log-likelihood of the data given a sample $\mathbf{z}_{0}$ drawn in latent space, and also called reconstruction loss, measuring how well we can reconstruct the data from the latent space.
The likelihood of the data $p(\mathbf{x} \mid \mathbf{z}_{0})$ maps back from latent space into data space and, hence, contains the decoder.
The posterior distribution $q(\mathbf{z}_{0} \mid \mathbf{x})$ specifies the first noised state given a data sample and includes the encoder, mapping from data space to latent space.
We will further discuss different options for the data log-likelihood in Appendix \ref{app:censored_dist}.

The second term is the Kullback-Leibler divergence between the distribution at $\tau=1$, the end of the diffusion process, and a prior distribution for the diffusion process.
Given the variance-preserving diffusion process from Eq. \eqref{eq:diff_process} and assuming a Gaussian prior distribution with mean $\boldsymbol{0}$ and covariance $\identity$, $p(\mathbf{z}_{1})=\gauss{\boldsymbol{0}}{\identity}$,
we obtain for the second term,
\begin{equation}
    \kl[\bigg]{q(\mathbf{z}_{1} \mid \mathbf{x})}{p(\mathbf{z}_{1})} = \frac12 \sum^{N_{\mathsf{latent}}}_{n=1} \Big [\sigma^{2}_{1} + (\alpha_{1}z_{x, n})^{2} - 1 - \log(\sigma^{2}_{1})\Big],
\end{equation}
over the $N_{\mathsf{latent}}$ dimensions of the latent space.

The last term is the diffusion loss, the only term that depends on the denoising NN $D_{\phi}(\mathbf{z}_{\tau}, \tau)$.
Given a sample in latent space $\mathbf{z}_{x}$ and a drawn pseudo-time $\tau$, we can calculate a noised sample $\mathbf{z}_{\tau}$ and the loss reads
\begin{equation}
    \mathcal{L}(\boldsymbol{\phi}, \mathbf{z}_{x}, \tau) = \Big(-\frac{d\gamma}{d\tau}\Big)\cdot e^{\gamma}\lVert \mathbf{z}_{x}- D_{\boldsymbol{\phi}}(\mathbf{z}_{\tau}, \tau)\rVert_{2}^{2}.\label{eq:diff_elbo_loss}
\end{equation}
Our loss function used to train the diffusion models, Eq. \ref{eq:diff_loss}, corresponds to this loss term by setting $w(\gamma)=1$.
As shown in \citet{kingma_understanding_2023}, using a monotonic weighting, like $w(\gamma)=\exp(-\frac{\gamma}{2})$, corresponds to the ELBO with data augmentation.
The only variable in Eq. \eqref{eq:diff_elbo_loss} are the weights and biases of the diffusion model and the noise scheduler that specifies $\gamma(\tau)$.
Using the observation that the noise scheduling can be seen as importance sampling, we use an adaptive noise scheduler as explained in Appendix \ref{app:noise_scheduler}.

In our experiments with diffusion models, we fix the end points of the noise scheduler to $\gamma(0) = \gamma_{min} = -15$ and $\gamma(1) = \gamma_{max} = 15$, and use a fixed or pre-trained encoder and decoder.
The first two terms of Eq. \eqref{eq:elbo} are then constant with respect to the trained diffusion model and can be neglected in the optimisation.
We obtain Eq. \eqref{eq:diff_elbo_loss} as the only loss to optimise the diffusion model.

Instead of directly predicting the state $\mathbf{z}_{x}$, we predict with our diffusion model a surrogate target $\mathbf{v}_{\tau} \coloneqq \alpha_{\tau}\boldsymbol{\epsilon} - \sigma_{\tau}\mathbf{z}_{x}$, as this tends to improve the convergence and the stability of the training \cite{salimans_progressive_2022}. 
We can recover the denoised state by setting $D_{\boldsymbol{\phi}}(\mathbf{z}_{\tau}, \tau) = \alpha_{\tau} \mathbf{z}_{x} - \sigma_{\tau} v_{\boldsymbol{\phi}}(\mathbf{z}_{\tau}, \tau)$.
Our loss function reads then
\begin{equation}
    \mathcal{L}(\boldsymbol{\phi}, \mathbf{z}_{\tau}, \tau) = w(\gamma) \cdot \Big(-\frac{d\gamma}{d\tau}\Big) \cdot (e^{-\gamma}+1)^{-1}\lVert\mathbf{v}_{\tau} - v_{\boldsymbol{\phi}}(\mathbf{z}_{\tau}, \tau)\rVert^2_2,
\end{equation}
which corresponds to the ELBO with data augmentation and $\mathbf{v}_{\tau}$-prediction.

To optimise the encoder and decoder together with the diffusion in an end-to-end training scheme \cite{vahdat_score-based_2021, shmakov_end--end_2023}, we can make use of Eq. \eqref{eq:elbo}.
However, pre-training the encoder and decoder simplifies and stratifies the training process.
To pre-train the encoder and decoder, we apply a variational autoencoder loss function,
\begin{subequations}
    \begin{align}
        \mathcal{L}(\bm{\theta}) = &-\expect_{q(\mathbf{z}_{x} \given \mathbf{x})}\Big[\log\big(p(\mathbf{x} \given \mathbf{z}_{x})\big)\Big]\tag{\ref{eq:recon_loss}}\\
        &+ \beta \kl[\Big]{q(\mathbf{z}_{x} \given \mathbf{x})}{p(\mathbf{z}_{x})}.\tag{\ref{eq:prior_loss}}
    \end{align}
\end{subequations}
Comparing Eq. \eqref{eq:elbo} with this pre-training loss, we can see that the reconstruction loss appears in both formulations, while the loss terms in latent/noised space are combined into a single loss, and the Eq. \eqref{eq:prior_loss} should prepare the latent space for its use in diffusion models.
The reconstruction loss is discussed more in detail in Appendix \ref{app:censored_dist}.

We train two different version of diffusion models:
\begin{itemize}
    \item The encoder and decoder are pre-trained with Eq. \eqref{eq:autoencoder_loss} as $\mathbf{z}_{x} = \mathrm{enc}(\mathbf{x})$ and $\hat{\mathbf{x}} = \mathrm{dec}(\mathbf{z}_{x})$, which results into a latent diffusion model.
    \item The encoder and decoder are fixed to the identity functions $\mathbf{z}_{x} = \mathbf{x}$ and $\hat{\mathbf{x}} = \mathbf{z}_{x}$, which leads to diffusion models in data space.
\end{itemize}
During their training, we draw a mini-batch of data samples $\mathbf{x} \drawn \mathcal{D}$, converted into latent space $\mathbf{z}_{x}$, and pseudo-times $\tau \in [0, 1]$.
We reduce the variance of the sampling in the pseudo-times by using stratified sampling as proposed in \citet{kingma_variational_2021}.
Afterwards, we estimate the diffusion loss, Eq. \eqref{eq:diff_loss}, for the current parameters $\phi$ of the diffusion model.
This loss is then used in Adam \cite{kingma_adam_2017} to make a gradient descent step.

The diffusion process can be seen as stochastic differential equation (SDE) that is integrated in pseudo-time \cite{song_maximum_2021}.
The reversion of this process is again a SDE \cite{anderson_reverse-time_1982} and we can find an ordinary differential equation (ODE) that has the same marginal distribution as the reverse SDE \cite{song_maximum_2021},
\begin{equation}
    d\widetilde{\mathbf{z}}_{\tau} = \Big[\mathbf{f}(\widetilde{\mathbf{z}}_{\tau}, \tau) - \frac12 g(\tau)^2 \mathbf{s}_{\boldsymbol{\phi}}(\widetilde{\mathbf{z}}_{\tau}, \tau) \Big]d\tau.\label{eq:diff_ode}
\end{equation}
For the variance-preserving diffusion process as defined in Eq. \eqref{eq:diff_process}, we obtain for the drift and diffusion term
\begin{align}
    \mathbf{f}(\widetilde{\mathbf{z}}_{\tau}, \tau) &= -\frac12\Big(\frac{d}{d\tau}\log(1+e^{-\gamma(\tau)})\Big)\widetilde{\mathbf{z}}_{\tau}\label{eq:ode_drift}\\
    g(\tau)^2 &= \frac{d}{d\tau}\log(1+e^{-\gamma(\tau)})\label{eq:ode_diffusion}.
\end{align}
Consequently, the only variable in the ODE, Eq. \eqref{eq:diff_ode}, is the score function $\mathbf{s}_{\boldsymbol{\phi}}(\widetilde{\mathbf{z}}_{\tau}, \tau)$ which we approximate with our trained NN,
\begin{equation}
    \mathbf{s}_{\boldsymbol{\phi}}(\widetilde{\mathbf{z}}_{\tau}, \tau) = -\Big(\widetilde{\mathbf{z}}_{\tau} + \frac{\alpha_{\tau}}{\sigma_{\tau}}v_{\boldsymbol{\phi}}(\widetilde{\mathbf{z}}_{\tau}, \tau)\Big)\label{eq:ode_score}.
\end{equation}
Given Eq. \eqref{eq:ode_drift}--\eqref{eq:ode_score}, we can integrate \eqref{eq:diff_ode} from $\tau=1$ to $\tau=0$ to find a solution and generate new samples; the initial conditions for the ODE are given by $\widetilde{\mathbf{z}}_{1} \drawn \gauss[]{\bm{0}}{\identity}$.
For the integration of the ODE, we use a second-order Heun sampler \cite{karras_elucidating_2022} with 20 integration steps.
The time steps for the integration are given by the inference noise scheduler as proposed by \citet{karras_elucidating_2022}.
We extend the noise scheduler to a wider range of signal-to-noise ratio by setting $\gamma_{min}=-15$ and $\gamma_{max}=15$ with truncation \cite{kingma_understanding_2023}.
Additionally, we adapt the sampling procedure: we apply the Heun procedure for all 20 integration steps, while we use a single Euler step to denoise the output from $\widetilde{\mathbf{z}}_{0}$ to $\widetilde{\mathbf{z}}_{x}$.
The integration hence includes 41 iterations with the trained neural network.
After obtaining $\widetilde{\mathbf{z}}_{x}$, we apply the decoder to map back into data space $\widetilde{\mathbf{x}} = \mathrm{dec}(\widetilde{\mathbf{z}}_{x})$ which gives us then the generated sample.

\subsection{Data log-likelihood with censored distributions}\label{app:censored_dist}

The autoencoder should map from latent space to data space by taking physical bounds into account, e.g. the non-negativity of the sea-ice thickness.
To achieve such bounds, we can clip the decoder output and set values below (over) the lower (upper) bound explicitly to the bound; similar to what the rectified linear unit (relu) activation function does for the negative bound.
Hence, all values below (above) the bound are projected to the bound.
In this clipping case, the output of the decoder is no longer the physical quantity itself but another latent variable, which is converted into the physical quantity in a post-processing step.
To differentiate between physical quantities and additional latent variable, we denote the decoder output as predicted latent variable $\widehat{\mathbf{y}} \in \mathbb{R}^{5 \times 512 \times 512}$  in the following.
We assume that this predicted latent variable is Gaussian distributed where the decoder predicts the mean $\bm{\mu} = \mathrm{dec}(\mathbf{z}_{x})$ and the standard deviation $\scale$ is spatially shared with one parameter per variable,
\begin{equation}
    \widehat{\mathbf{y}} \sim \gauss[\big]{\widehat{\bm{\mu}}}{\scale^2 \identity}.\label{eq:gaussian_latent}
\end{equation}
Caused by the diagonal covariance, we make out of the multivariate data log-likelihood a univariate estimation, which we can simply sum over the variables and grid points.
Given this Gaussian assumption, the data log-likelihood would read for $K$ variables and $L$ grid points
\begin{equation}
    -\log p(\mathbf{x} \mid \mathbf{z}_{x}) = \frac{1}{2}\sum^{K}_{k=1}\sum^{L}_{l=1}\frac{(x_{k, l}-\widehat{\mu}_{k, l})^2}{(s_{k})^2} + \log\big((s_{k})^2\big) + \log(2 \cdot \pi).\label{eq:gaussian_nll}
\end{equation}

The log-likelihood is proportional to a mean-squared error (MSE) weighted by the scale parameter.
Although the MSE is often employed to train an autoencoder, we make the assumption that the decoder output is the physical quantity itself and we neglect the clipping in the training of the autoencoder.
Consequently, if we clip the output into physical bounds, we make a wrong assumption and introduce a bias into the decoder and, thus, the latent space.
This bias depends on the number of cases that the lower (upper) bound is reached and is difficult to quantify.
To explicitly bake the clipping into the cost function and treating the decoder output as latent variable, we censor the assumed distribution.

To optimise the autoencoder, we derive the data log-likelihood where variables are lower and upper bounded with $\lowerbound$ as lower and $\upperbound$ as upper bound, e.g., $x_{\mathsf{L}} = 0$ as lower and $x_{\mathsf{U}} = 1$ as upper bound for the sea-ice concentration.
For variables with only one bound, the case of the missing bound can be simply omitted.

The predicted latent variable for the $k$-th variable and the $l$-th grid point is clipped into the physical bounds by
\begin{equation}
    \widehat{\mathbf{x}}_{k, l} = \begin{cases}
        x_{\mathsf{L}, k}, & \text{if } \widehat{\mu}_{k, l} \leq x_{\mathsf{L}, k}\\
        x_{\mathsf{U}, k}, & \text{if } \widehat{\mu}_{k, l} \geq x_{\mathsf{U}, k}\\
        \widehat{\mu}_{k, l}, & \text{otherwise}.
    \end{cases}\label{eq:clipping}
\end{equation}
This clipping operation is non-invertible such that we cannot recover the true latent variable $y_{k, l}$ from observing the physical quantity $x_{k, l}$.
However, we know that if $x_{k, l}=x_{\mathsf{L}, k}$ or $x_{k, l}=x_{\mathsf{U}, k}$, the true latent variable was clipped.
In such cases, rationally, the data log-likelihood hence should increase the probability that the decoder output is clipped.

In the following, we denote the cumulative distribution function (CDF) of the normal Gaussian distribution as $\cdf$ and its probability density function (PDF) as $\pdf$. 
Given the assumed Gaussian distribution of the latent variables, Eq. \eqref{eq:gaussian_latent}, the physical quantity is presumably distributed by a censored Gaussian distribution with the following PDF for the $k$-th variable and $l$-th grid point,
\begin{equation}
    \pdist{x_{k, l} \given \mathbf{z}_{x}} = \begin{cases}
        \cdf(\frac{x_{\mathsf{L}, k}-\widehat{\mu}_{k, l}}{s_k}), & \text{if } x_{k, l} = x_{\mathsf{L}, k},\\
        \cdf(\frac{\widehat{\mu}_{k, l}-x_{\mathsf{U}, k}}{s_k}), & \text{if } x_{k, l} = x_{\mathsf{U}, k},\\
        \frac{1}{s_k}\pdf(\frac{x_{k, l}-\widehat{\mu}_{k, l}}{s_k}), & \text{if } x_{\mathsf{L}, k} < x_{k, l} < x_{\mathsf{U}, k},\\
        0, & \text{otherwise}.
    \end{cases}\label{eq:censored_gaussian}
\end{equation}
Different from a truncated Gaussian distribution, Eq. \eqref{eq:censored_gaussian} gives values at the bounds a probability larger than zero, and all the density of the latent variable exceeding the bounds is folded to the bounds.

Using the PDF of the censored distribution from Eq. \eqref{eq:censored_gaussian} for the data log-likelihood, we obtain
\begin{equation}
    \begin{split}
    -\log p(\mathbf{x} \mid \mathbf{z}_{x}) = \sum^{K}_{k=1}\sum^{L}_{l=1} \phantom{+} &I(x_{k, l} = x_{\mathsf{L}, k})\,\log\bigg(\cdf\bigg(\frac{x_{\mathsf{L}, k}-\widehat{\mu}_{k, l}}{s_k}\bigg)\bigg)\\
    +& I(x_{k, l} = x_{\mathsf{U}, k})\,\log\bigg(\cdf\bigg(\frac{\widehat{\mu}_{k, l}-x_{\mathsf{U}, k}}{s_k}\bigg)\bigg)\\
    +& I(x_{\mathsf{L}, k} < x_{k, l} < x_{\mathsf{U}, k})\,\log\bigg(\frac{1}{s_k}\pdf\bigg(\frac{x_{k, l}-\widehat{\mu}_{k, l}}{s_k}\bigg)\bigg)
    \end{split}\label{eq:loss_censored}
\end{equation}
as cost function, where $I(\cdot)$ is the indicator function.
The first two parts of the cost function are the bounded cases and correspond to a Gaussian classification model, using as probability the log-CDFs at the bounds.
The last part is the log-likelihood of a Gaussian distribution as in Eq. \eqref{eq:gaussian_nll} and includes the weighted MSE.
The here derived log-likelihood thus combines a regression with a classification and is the negative log-likelihood of a type I Tobit model \citep{tobin_estimation_1958}.
Whereas the optimisation of Eq. \eqref{eq:loss_censored} seems more complicated than the optimisation of a log-likelihood from a Gaussian distribution, we have experienced no optimisation issues with variants of gradients descent.

Caused by the regression-classification mixture, the decoder output specifies not only the variable of interest but combined with the scale also the probability that the output will be clipped.
This censored distribution approach works because we neglect correlations between variables and grid points and convert the multivariate into a univariate prediction problem.
Nevertheless, we are free to chose any neural network architecture to predict the mean of the distribution in Eq. \eqref{eq:gaussian_latent}.
Consequently, by e.g. using a fully convolutional decoder, we can still represent correlations in the decoder output.

\subsection{Adaptive noise scheduler}\label{app:noise_scheduler}

The noise scheduler maps a given pseudo-time $\tau \in [0, 1]$ to a log-signal-to-noise ratio $\gamma(\tau)$ and determines how much time is spent at a given ratio.
In the manuscript, we use different schedulers during training and generation.
While we generate the data with a fixed scheduler as proposed by \citet{karras_elucidating_2022}, we make use of an adaptive scheduler \cite{kingma_understanding_2023} during training.
In the following, we briefly introduce the principles of the adaptive scheduler, and we refer to \citet{kingma_understanding_2023} for a longer treatment.

As shortly shown in Appendix \ref{app:add_methods_diffusion}, the loss function of the diffusion model optimises the evidence lower bound (ELBO).
Consequently, we can extend the variational parameters by the parameters of the training noise scheduler to further tighten the bound \cite{kingma_variational_2021} which gives us the optimal scheduler in the ELBO sense.
Extending on this idea, the distribution given after transforming a drawn pseudo-time into $\gamma$ can be seen as importance sampling distribution $p(\gamma)$  \cite{kingma_understanding_2023}, and we can write
\begin{equation}
    -\frac{d\gamma}{d\tau} = \frac{1}{p(\gamma)}.
\end{equation}
Interpreting the output of the noise scheduler as a drawn from an importance sampling distribution allows us to formulate the optimal noise scheduler which should fulfil
\begin{equation}
        p(\gamma) \propto \expect_{\mathbf{x} \drawn \dataset, \epsilon \drawn \gauss{\bm{0}}{\identity}} \Big[ w(\gamma) \cdot (e^{-\gamma}+1)^{-1}\lVert\mathbf{v}_{\gamma} - v_{\boldsymbol{\phi}}(\mathbf{z}_{\gamma}, \gamma)\rVert^2_2 \Big],\label{eq:importance_sampling}
\end{equation}
where we made a change-of-variables from $\tau$ to $\gamma$ to signify the dependency on $\gamma$.

As Eq. \eqref{eq:importance_sampling} is infeasible, we approximate the optimal importance sampling distribution.
Using 100 equal-distant bins between $\gamma_{\mathrm{min}} = -15$ and $\gamma_{\mathrm{max}} = 15$, we keep track of an exponential moving average of Eq. \eqref{eq:importance_sampling} for each bin.
The density is constant within each bin, which allows us to estimate an empirical cumulative distribution function from $\gamma_{\mathrm{max}}$ to $\gamma_{\mathrm{min}}$.
The training noise scheduler is then given as empirical quantile function mapping from $[0, 1]$ to $[\gamma_{\mathrm{max}}, \gamma_{\mathrm{min}}]$.
In practice, this adaptive noise scheduler reduces the number of needed hyperparameters for the diffusion model and improves its convergence \cite{kingma_understanding_2023}.
It was recently similarly used in \citet{finn_generative_2024}.

\section{Architectures}\label{app:architectures}

In the following, we will describe our neural network architectures.
The autoencoder is purely based on a convolutional architecture, while the diffusion models are based on a mixture between a convolutional U-Net \cite{ronneberger_u-net_2015} and a vision transformer (ViT, \citet{dosovitskiy_image_2021}), termed UVit \cite{hoogeboom_simple_2023}.

\subsection{Masked convolutions}

As visible in Fig. \ref{fig:samples}, sea ice only exists on grid points covered by ocean while grid points with land are omitted.
To take the masking of land pixels into account, we apply convolutional operations as inspired by partial convolutions \cite{liu_image_2018, kadow_artificial_2020, durand_data-driven_2024}.

Before each convolutional operation, the data is multiplied by the mask $\mathbf{m} \in [0, 1]$, where valid grid points are $1$.
The masking sets all grid points with land to zero such that they act like zero padding and interactions with land grid points are deactivated.
Consequently, grid points near land exchange less information with the surrounding grid points which in fact imitates how land masking is implemented in numerical models.

\subsection{Autoencoder}\label{app:ae_architecture}

The autoencoder maps from $\mathbf{x} \in \mathbb{R}^{5 \times 512 \times 512}$ to $\mathbf{z}_{z} \in \mathbb{R}^{16 \times 64 \times 64}$ and vice-versa.
Its encoder includes a series of downsampling operations with ConvNeXt blocks, while the decoder combines upsampling operations with ConvNeXt blocks.

The main computing block are \textbf{ConvNeXt blocks} \cite{liu_convnet_2022}: a wide masked convolution with a $7 \times 7$ kernel extracts channel-wise spatial information.
This spatial information is normalised with layer normalisation \cite{ba_layer_2016}.
Afterwards, a multi-layered perceptron (MLP) with a Gaussian error linear unit (Gelu, \citet{hendrycks_gaussian_2016}) activation mixes up the normalised information, before the output gets added again to the input of the block in a residual connection.
Throughout the ConvNeXt block, the number of channels remain the same.

The \textbf{downsampling} in the encoder combines layer normalisation with a masked convolution and a $2\times2$ kernel and a stride of $2$, which also doubles the number of channels.
To downsample the mask, we apply max pooling, i.e. if there is an ocean grid point in a $2\times2$ window, the output grid point is set to $1$. 
This increases the number of ocean grid points compared to a strided downsampling.

In the \textbf{upsampling} of the decoder, a layer normalisation is followed by a nearest neighbour interpolation, which doubles the number of grid points in $y$- and $x$-direction.
Afterwards, a masked convolution with a $3\times3$ kernel smooths the interpolated fields and halves the number of channels.
This combination of interpolation with convolution results into less checkerboard effects compared to a transposed convolution \cite{odena_deconvolution_2016}.

The architectures for the encoder and decoder read (the number in brackets represents the number of channels):
\begin{itemize}
    \item{\textbf{Encoder:} Input(5) $\rightarrow$ Point-wise convolution (64) $\rightarrow$ ConvNeXt (64) $\rightarrow$ ConvNeXt (64) $\rightarrow$ Downsampling (128) $\rightarrow$ ConvNeXt (128) $\rightarrow$ ConvNeXt (128) $\rightarrow$ Downsampling (256) $\rightarrow$ ConvNeXt (256) $\rightarrow$ ConvNeXt (256) $\rightarrow$ Downsampling (512) $\rightarrow$ ConvNeXt (512) $\rightarrow$ ConvNeXt (512) $\rightarrow$ Layer normalisation (512) $\rightarrow$ rectified linear unit (relu, 512) $\rightarrow$ Point-wise convolution (32)}
    \item{\textbf{Decoder:} Input(16) $\rightarrow$ Point-wise convolution (512) $\rightarrow$ ConvNeXt (512) $\rightarrow$ ConvNeXt (512) $\rightarrow$ Upsampling (256) $\rightarrow$ ConvNeXt (256) $\rightarrow$ ConvNeXt (256) $\rightarrow$ Upsampling (128) $\rightarrow$ ConvNeXt (128) $\rightarrow$ ConvNeXt (128) $\rightarrow$ rightarrow (64) $\rightarrow$ ConvNeXt (64) $\rightarrow$ ConvNeXt (64) $\rightarrow$ Layer normalisation (64) $\rightarrow$ relu (64) $\rightarrow$ Point-wise convolution (5)}.
\end{itemize}
Note, the encoder outputs the mean and standard deviation in latent space, while the decoder gets only a latent sample where the mean and standard deviation are combined.
The relu activation before the last point-wise convolution helps to improve the representation of continuous-discrete sea-ice processes.

In total, the encoder has $2.2\times10^{6}$ parameters and the decoder has $3.1\times10^{6}$ parameters.

\subsection{Diffusion architecture}\label{app:diff_architecture}

The diffusion models modify the UViT architectures as introduced in \citet{hoogeboom_simple_2023}.
The encoding and decoding part of the architecture are implemented with convolutional operations, similar to how we implemented the autoencoder.
In the bottleneck, the architecture uses a vision transformer \cite{dosovitskiy_image_2021}.

In addition to the latent sample $\mathbf{z}_{\tau}$, the diffusion models get the pseudo-time $\tau$ as input.
This pseudo-time is handled as conditioning information and embedded with a sinusoidal embedding \cite{vaswani_attention_2017} to extract 512 Fourier features.
The embedding is followed by a MLP which reduces the number of features to 256 and where we apply a Gelu for feature activation.
These extracted features are fed into the adaptive layer normalisation layers \cite{perez_film_2017} with conditioned affine parameters.

The main computing blocks of the encoding and decoding part are again \textbf{ConvNeXt} blocks (see also Appendix \ref{app:ae_architecture}), with an additional conditioning of the residual connections on the pseudo-time features.
The \textbf{downsampling} in the encoding part remains the same as for the encoder.
For both blocks, we simply replace layer normalisation by adaptive layer normalisation conditioned on the pseudo-time.

The \textbf{upsampling} is similar to the upsampling for the decoder (see also Appendix \ref{app:ae_architecture}).
However, we have also skip connections, transferring information from before the downsampling to the upsampling at the same level.
These features are concatenated with the interpolated features before the convolution of the upsampling is applied.
Consequently, the convolution has $1.5\times$ more input channels than the convolution in the upsampling of the decoder.
Additionally, we again replace layer normalisation by adaptive layer normalisation conditioned on the pseudo-time.

The \textbf{transformer} blocks \cite{vaswani_attention_2017} closely resemble the default implementation of ViT transformer blocks \cite{dosovitskiy_image_2021}.
A self-attention block with 8 heads is followed by a MLP.
Throughout the transformer block, the number of channels remain the same, even in the multi-layered perceptron.
We apply adaptive layer normalisation at the beginning of each residual layer in the self-attention block and MLP \cite{xiong_layer_2020} and additionally conditioned the residual connection on the pseudo-time.
We remove land grid points before applying the transformer and add them afterwards again.

The architectures for the diffusion model in data space and in latent space read then (the number in brackets represents the number of channels):
\begin{itemize}
    \item{
        \textbf{Data space:} Input(5) $\rightarrow$ Point-wise convolution (64) 
        $\rightarrow$ ConvNeXt (64) $\rightarrow$ ConvNeXt (64) $\rightarrow$ Downsampling (64)
        $\rightarrow$ ConvNeXt (64) $\rightarrow$ ConvNeXt (64) $\rightarrow$ Downsampling (64)
        $\rightarrow$ ConvNeXt (64) $\rightarrow$ ConvNeXt (64) $\rightarrow$ Downsampling (64)
        $\rightarrow$ ConvNeXt (64) $\rightarrow$ ConvNeXt (64) $\rightarrow$ Downsampling (64)
        $\rightarrow$ ConvNeXt (64) $\rightarrow$ ConvNeXt (64) $\rightarrow$ Downsampling (128)
        $\rightarrow$ ConvNeXt (128) $\rightarrow$ ConvNeXt (128) $\rightarrow$ Downsampling (256)
        $\rightarrow$ Transformer (256) $\rightarrow$ Transformer (256) $\rightarrow$ Transformer (256) $\rightarrow$ Transformer (256) $\rightarrow$ Transformer (256) $\rightarrow$ Transformer (256) $\rightarrow$ Transformer (256) $\rightarrow$ Transformer (256)
        $\rightarrow$ Upsampling (128) $\rightarrow$ ConvNeXt (128) $\rightarrow$ ConvNeXt (128)
        $\rightarrow$ Upsampling (64) $\rightarrow$ ConvNeXt (64) $\rightarrow$ ConvNeXt (64)
        $\rightarrow$ Upsampling (64) $\rightarrow$ ConvNeXt (64) $\rightarrow$ ConvNeXt (64)
        $\rightarrow$ Upsampling (64) $\rightarrow$ ConvNeXt (64) $\rightarrow$ ConvNeXt (64)
        $\rightarrow$ Upsampling (64) $\rightarrow$ ConvNeXt (64) $\rightarrow$ ConvNeXt (64)
        $\rightarrow$ Upsampling (64) $\rightarrow$ ConvNeXt (64) $\rightarrow$ ConvNeXt (64)
        $\rightarrow$ Layer normalisation (64) $\rightarrow$ relu (64) $\rightarrow$ Point-wise convolution (5)
    }
    \item{
        \textbf{Latent space:} Input(16) $\rightarrow$ Point-wise convolution (64)
        $\rightarrow$ ConvNeXt (64) $\rightarrow$ ConvNeXt (64) $\rightarrow$ Downsampling (128)
        $\rightarrow$ ConvNeXt (128) $\rightarrow$ ConvNeXt (128) $\rightarrow$ Downsampling (256)
        $\rightarrow$ ConvNeXt (256) $\rightarrow$ ConvNeXt (256) $\rightarrow$ Downsampling (512)
        $\rightarrow$ Transformer (512) $\rightarrow$ Transformer (512) $\rightarrow$ Transformer (512) $\rightarrow$ Transformer (512) $\rightarrow$ Transformer (512) $\rightarrow$ Transformer (512) $\rightarrow$ Transformer (512) $\rightarrow$ Transformer (512)
        $\rightarrow$ Upsampling (256) $\rightarrow$ ConvNeXt (256) $\rightarrow$ ConvNeXt (256)
        $\rightarrow$ Upsampling (128) $\rightarrow$ ConvNeXt (128) $\rightarrow$ ConvNeXt (128)
        $\rightarrow$ Upsampling (64) $\rightarrow$ ConvNeXt (64) $\rightarrow$ ConvNeXt (64)
        $\rightarrow$ Layer normalisation (64) $\rightarrow$ relu (64) $\rightarrow$ Point-wise convolution (5).
    }
\end{itemize}
Note, caused by the high dimensionality of the data, the diffusion model in data space has a high memory consumption, and we kept $64$ channels for longer than in the latent diffusion model.
The diffusion model has $15.3\times10^{6}$ parameters and the latent diffusion model $13.4\times10^{6}$ parameters.

\section{Experiments}\label{app:exp_details}

In our experiments, we train autoencoders and diffusion models with the architectures as described in Appendix \ref{app:architectures}.
As we train different types of models, we designed our experiments with a unified setup to achieve a fair comparison.

We use sea-ice simulation data from the state-of-the-art sea-ice model neXtSIM \cite{rampal_nextsim_2016, olason_new_2022}.
This sea-ice model uses an advanced brittle rheology \cite{girard_new_2011, dansereau_maxwell_2016, olason_new_2022} which introduces a \textit{damage} variable to represent subgrid-scale dynamics.
This way, the model can represent sea-ice dynamics at the mesoscale resolution $\simeq 10\,\text{km}$ as they are observed with satellites and buoys \cite{rampal_multi-fractal_2019, olason_statistical_2021, bouchat_sea_2022}.
In the here used simulations \cite{boutin_arctic_2023}, neXtSIM has been coupled to the ocean component OPA of the NEMO modelling framework \cite{madec_nemo_2008}.
NeXtSIM runs on a Lagrangian mesh with adaptive remeshing, while OPA uses a curvilinear mesh, here in the regional CREG025 configuration \cite{talandier_creg025l75-nemo_r360_2021}.
Both models are run at a similar resolution of $\frac{1}{4}^{\circ} \simeq 12 \,\text{km}$.
The output of neXtSIM is conservatively interpolated to the curvilinear OPA mesh.
NeXtSIM is driven by ERA5 atmospheric forcings \cite{hersbach_era5_2020}.
Since we use unconditional generation of fields, we neglect model output from OPA and the atmospheric forcing for the training of the neural networks.

We target to generate samples for five different variables, the sea-ice thickness, the sea-ice concentration, the two sea-ice velocities in $x$- and $y$-direction, and the sea-ice damage, a special variable introduced for brittle rheologies which represents the subgrid-scale fragmentation of sea ice.
The model output for all variables is a six-hourly average, while the damage corresponds to a six-hourly instantaneous output.
In correspondence with \citet{boutin_arctic_2023}, we use data from 2000--2018, omitting the first five years (1995-1999) as spin-up.
The first 14 years (2000-2014, 21916 samples) are used as training dataset, 2015 as validation dataset (1460 samples), and 2016--2018 for testing (4383 samples).
All datasets are normalised by a per-variable global mean and standard deviation estimated over the training dataset.

Given these datasets, we optimise all models with Adam \cite{kingma_adam_2017} with a batch size of 16.
The learning rate is linearly increased from $1\times10^{-6}$ to $2\times10^{-4}$ within $5 \times 10^{3}$ iterations.
Afterwards a cosine scheduler is used to decrease the learning rate again up to $1\times10^{-6}$ after $10^{5}$ iterations, where the training is stopped.
All models are trained without weight decay and other regularisation techniques like dropout, and we select the models that achieve the lowest validation loss.
We train the models with mixed precision in \textit{bfloat16} and evaluate in single precision.
All models are trained on either a NVIDIA RTX5000 GPU (24 GB) or a NVIDIA RTX6000 GPU (48 GB); to train on the RTX5000, we use a batch size of 8 and accumulate the gradient of two iterations.
All models are implemented in PyTorch \cite{paszke_pytorch_2019} with PyTorch lightning \cite{falcon_pytorchlightning_2020} and Hydra \cite{yadan_hydra_2019}.
The code is available under \url{https://github.com/cerea-daml/ldm_nextsim}.

To train the latent diffusion models (LDMs), we use the mean prediction of the pre-trained encoder as deterministic mapping from pixel into latent space.
Before training of the LDMs, we normalise the latent space by a per-channel global mean and standard deviation estimated over the training dataset.
Although the LDM needs much less memory than the autoencoder and pixel diffusion model during training, we stick to a batch size of $16$ for a fair comparison.
All diffusion models use the same fixed log-signal-to-ratio bounds $\gamma_{\mathrm{min}} = -15$ and $\gamma_{\mathrm{max}} = 15$ during training and sampling.
We perform the sampling with a second-order Heun sampler in 20 integration steps and the EDM sampling noise scheduler \cite{karras_elucidating_2022}, extended to our bounds by truncation \cite{kingma_understanding_2023}.

Both, the reconstructions and the generated data are compared to the testing dataset.
For the reconstructions, we perform a one-to-one comparison, while we evaluate the statistics of our generated data.
In total, we hence generate $4383$ samples with the diffusion models.
Although a smaller number than the $50000$ samples normally used in validating image diffusion models, it results into a fair comparison to the testing dataset.
In Appendix \ref{app:results_dataset_size}, we demonstrate that the dataset size is big enough to see differences in the models.

\section{Metrics}\label{app:metrics}

In the following we introduce the metrics to evaluate our neural networks.
To evaluate the autoencoders, we apply point-wise measures, while we quantify the quality of the generated fields in a transformed space.
The reconstructed samples $\widetilde{\mathbf{x}} \in \mathbb{R}^{N \times K\times L}$, and generated samples $\widehat{\mathbf{x}} \in \mathbb{R}^{N \times K\times L}$ are compared to the samples in the testing dataset $\mathbf{x} \in \mathbb{R}^{N \times K\times L}$ for $N=4383$ samples, $K=5$ variables, and $L=127152$ grid points without land.

\subsection{Root-mean-squared error}

To estimate one averaged root-mean-squared error (RMSE), we normalise the RMSE by the per-variable standard deviation $\sigma_{k}$, globally calculated for the $k$-th variable over the full training dataset.
The normalised RMSE is then calculated as
\begin{equation}
    \mathrm{RMSE}(\mathbf{x}, \widetilde{\mathbf{x}}) = \sqrt{\frac{1}{N \cdot K \cdot L} \sum_{n=1}^{N}\sum_{k=1}^{K}\sum_{l=1}^{L} \frac{(x_{n, k, l} -\widetilde{x}_{n, k, l})^2}{(\sigma_{k})^2}}.
\end{equation}
The lowest (best) possible RMSE is $0$ and its maximum value is unbounded.

\subsection{Structural similarity index measure}

The structural similarity index measure ($\mathrm{SSIM}$) takes information from a window into account and is hence a more fuzzy verification metric than the MAE.
The SSIM for the $n$-th sample, the $k$-th variable, and the $l$-th is given as
\begin{equation}
    \mathrm{SSIM}(x_{n, k, l},y_{n, k, l}) = \frac{\Big(2\mu^{(x)}_{n, k, l}\mu^{(y)}_{n, k, l}\Big)\Big(2\sigma^{(xy)}_{n, k, l} + c_{2, k}\Big)}{\Big((\mu^{(x)}_{n, k, l})^2 + (\mu^{(y)}_{n, k, l})^2 + c_{1, k}\Big)\Big((\sigma^{(x)}_{n, k, l})^2 + (\sigma^{(y)}_{n, k, l})^2 + c_{2, k}\Big)}.\label{eq:single_ssim}
\end{equation}
The means $\mu^{(x)}_{n, k, l}$ and $\mu^{(y)}_{n, k, l}$, variances $\sigma^{(x)}_{n, k, l}$ and $\sigma^{(y)}_{n, k, l}$, and covariance $\sigma^{(xy)}_{n, k, l}$ between $x$ and $y$, respectively are estimated in the given $7\times7$ window for each sample, variable, and grid point.
The stabilisation values $c_{1, k}=(0.01 \cdot r_{k})^2$ and $c_{2}=(0.03 \cdot r_{k})^2$ depend on the dynamic range for the $k$ variable with $r_{k} = max(\mathbf{x}_{k}) - min(\mathbf{x}_{k})$ as the difference between the maximum and minimum in the training dataset.

The global score for the SSIM is the average of Eq. \eqref{eq:single_ssim} across all samples, variables, and grid points
\begin{equation}
    \mathrm{SSIM}(\mathbf{x}, \widetilde{\mathbf{x}}) = \frac{1}{N \cdot K \cdot L} \sum_{n=1}^{N}\sum_{k=1}^{K}\sum_{l=1}^{L} \mathrm{SSIM}(x_{n, k, l},\widetilde{x}_{n, k, l}).
\end{equation}
The highest possible SSIM is $1$ and the lowest $0$.

\subsection{Sea-ice extent accuracy}

By applying diffusion models in a latent space, we can incorporate prior knowledge into the autoencoder which maps to and from the latent space.
To determine the need of incorporating this knowledge into the autoencoder, we evaluate the accuracy of reconstructing where the sea ice is.
While the accuracy might be unneeded for data generation, it can be a necessity for different tasks like surrogate modelling.

In neXtSIM, sea-ice processes are activated or deactivated based on the sea-ice thickness, and we estimate the sea-ice extent based on the sea-ice thickness.
Choosing a small threshold value of $SIT_{\mathsf{min}} = 0.01\,\text{m}$, we classify values above this threshold as sea ice and below as no ice (similar results can be achieved by using the sea-ice concentration).
Based on this classification, we estimate the average accuracy across all samples and grid points.
The accuracy is the sum of the true positive and true negative divided by the total number of cases.
This accuracy is the same as $1-\mathrm{IIEE}$, where $\mathrm{IIEE}$ is the integrated ice edge error \cite{goessling_predictability_2016} estimated based on the sea-ice thickness instead of the sea-ice concentration.
The highest possible accuracy is $1$ and the lowest $0$.

\subsection{Fréchet distance in latent space}

If we generate data, we have no one-to-one correspondence between samples from the testing dataset and generated samples.
To evaluate the generated statistics, we employ a latent space spanned by an independently trained variational autoencoder.
The variational autoencoder follows the general structure of our trained autoencoders with $\beta=1$.
As the evidence lower bound is maximised, we can expect that this variational autoencoder encodes spatial and semantic information into the latent space.
Additionally, the KL-divergence $\kl[\Big]{q(\mathbf{z}_{x} \given \mathbf{x})}{\gauss[\big]{\mathbf{0}}{\identity}}$ regularises the latent space towards an isotropic Gaussian distribution.
Hence, we assume that the latent space is distributed with an isotropic Gaussian.

To emulate how image generators are validated with the Fréchet Inception distance (FID, \citet{heusel_gans_2017}), we employ the Fréchet distance in latent space and call this Fréchet autoencoder distance ($\mathrm{FAED}$).
Specifically, we use the mean prediction of the encoding part $\mathrm{enc}(\mathbf{x})$ as deterministic mapping into latent space.
The Fréchet autoencoder distance reads then
\begin{equation}
    \mathrm{FAED}(\mathbf{x}, \widetilde{\mathbf{x}}) = \lVert \mu(\mathrm{enc}(\mathbf{x})) - \mu(\mathrm{enc}(\widetilde{\mathbf{x}}))\rVert_{2}^2 + \mathrm{Tr}\Bigg(\Sigma(\mathrm{enc}(\mathbf{x}))+\Sigma(\mathrm{enc}(\widetilde{\mathbf{x}}))-2\Big(\Sigma(\mathrm{enc}(\mathbf{x}))\cdot\Sigma(\mathrm{enc}(\widetilde{\mathbf{x}}))\Big)^{\frac12} \Bigg),
\end{equation}
where $\mu(\cdot)$ is the point-wise sample mean, $\Sigma(\cdot)$ the sample covariance, $\mathrm{Tr}(\cdot)$ the trace of a matrix, and $(\cdot)^\frac12$ the matrix square-root.
The isotropic Gaussian assumption in latent space allows us to further simplify the estimated covariances matrix to variances, and the Fréchet autoencoder distance reduces to
\begin{equation}
        \mathrm{FAED}(\mathbf{x}, \widetilde{\mathbf{x}}) = \lVert \mu(\mathrm{enc}(\mathbf{x})) - \mu(\mathrm{enc}(\widetilde{\mathbf{x}}))\rVert_{2}^2 + \lVert \sigma(\mathrm{enc}(\mathbf{x})) - \sigma(\mathrm{enc}(\widetilde{\mathbf{x}}))\rVert_{2}^2,
\end{equation}
with $\sigma(\cdot)$ as point-wise standard deviation.
The minimum (best) value of the Fréchet autoencoder distance is $0$ while its maximum value is unbounded.

\subsection{Root-mean-squared error of the sea-ice extent}

To evaluate the climatological consistency of the generated samples for the sea-ice extent, we can estimate the probability $\widetilde{p}_{l}$ that the $l$-th grid point is covered by sea ice.
To classify for the $n$-th sample if a grid point is covered, we again apply the threshold of $\mathrm{SIT}_{\mathrm{min}} = 0.01\,\text{m}$.
The probability is given as
\begin{equation}
    \widetilde{p}_{l} = \frac{1}{N}\sum^{N}_{n=1}\mathrm{I}(\widetilde{\mathrm{SIT}}_{n, l}\geq\mathrm{SIT}_{\mathrm{min}})
\end{equation}
with $\mathrm{I}(\cdot)$ as indicator function and $\widetilde{\mathrm{SIT}}_{n, l}$ the sea-ice thickness of the $n$-th generated sample for the $l$-th grid point.

The root-mean-squared error (RMSE) in the sea-ice extent reads then
\begin{equation}
    \mathrm{RMSE}_{\mathrm{SIE}}(\mathbf{x}, \widetilde{\mathbf{x}}) = \sqrt{\frac{1}{L}\sum^{L}_{l=1} (p_{l}-\widetilde{p}_{l})^2}
\end{equation}
with $p_{l}$ as probability in the testing dataset.
The minimum (best) RMSE is $0$, while the worst possible RMSE is $1$.

\section{Additional results}

In the Appendix, we present additional results that signify the results we have found.
We will show how the hyperparameters influence the autoencoder reconstruction, how censoring helps to improve the sea-ice extent representation, and how the diffusion model generalises across experiments.

\subsection{Ablation for autoencoder}\label{app:results_ablation_ae}

In our formulation for the pre-training of the autoencoder, the autoencoder has several hyperparameters.
One of the them is the number of channels $N_{\text{latent}}$ in latent space.
The more channels, the more information can be stored in the latent space.
However, more channels can make the diffusion model more difficult to train.
Another hyperparameter is the factor $\beta$ which determines the strength of the regularisation in latent space.
The larger the factor, the smoother the latent space, but also the higher the semantic compression therein.
Throughout the main manuscript, we have made the decision of $N_{\text{latent}}=16$ and $\beta=10^{-3}$ to strike the right balance.
In Tab. \ref{tab:app_ablation_ae}, we show an ablation study on these factors.
Different from the experiments in the main manuscript, these experiments were performed with NVIDIA A100 GPUs on the Jean-Zay supercomputing facilities, provided by GENCI.

\begin{table}[ht]
\caption{
    Evaluation of the autoencoders for reconstructing the testing dataset with the normalised root-mean-squared error (RMSE), the structural similarity index measure (SSIM), and the accuracy in the sea-ice extent ($\text{ACC}_{\text{SIE}}$).
    $N_{\text{latent}}$ is the number of channels in the latent space and $\beta$ is the regularisation factor.
    The two rows marked in bold are used in the main manuscript.
    Note, the numbers can be different than in Tab. \ref{tab:autoencoder} as these autoencoders are trained independently on another computing system.
}
\label{tab:app_ablation_ae}
\begin{center}
\begin{small}
\begin{sc}
\begin{tabular}{rrlrrr}
    \toprule
    $N_{\text{latent}}$ & $\beta$ & & RMSE & SSIM & $\text{ACC}_{\text{SIE}}$ \\
    \midrule
    8 & $10^{-3}$ & -- & 0.094 & 0.916 & 0.982 \\
    \textbf{16} & $\mathbf{10^{-3}}$ & \textbf{--} & \textbf{0.076} & \textbf{0.937} & \textbf{0.981} \\
    32 & $10^{-3}$ & -- & 0.061 & 0.953 & 0.984 \\
    16 & $10^{-2}$ & -- & 0.077 & 0.935 & 0.971 \\
    16 & $10^{-4}$ & -- & 0.075 & 0.939 & 0.984 \\
    \textbf{16} & $\mathbf{10^{-3}}$ & \textbf{censored} & \textbf{0.078} & \textbf{0.942} & \textbf{0.992} \\
    \bottomrule
\end{tabular}
\end{sc}
\end{small}
\end{center}
\end{table}

With our chosen hyperparameters, the autoencoder results into a $16.4$-fold compression, which is lower than the $20$-fold compression from data dimensions only as the dimensionality of the masking is also reduced.
Altering this compression rate by changing the number of channels in latent space has a larger impact on the RMSE and the SSIM than the regularisation factor.

\begin{figure}[ht]
\begin{center}
\centerline{\includegraphics[width=0.48\textwidth]{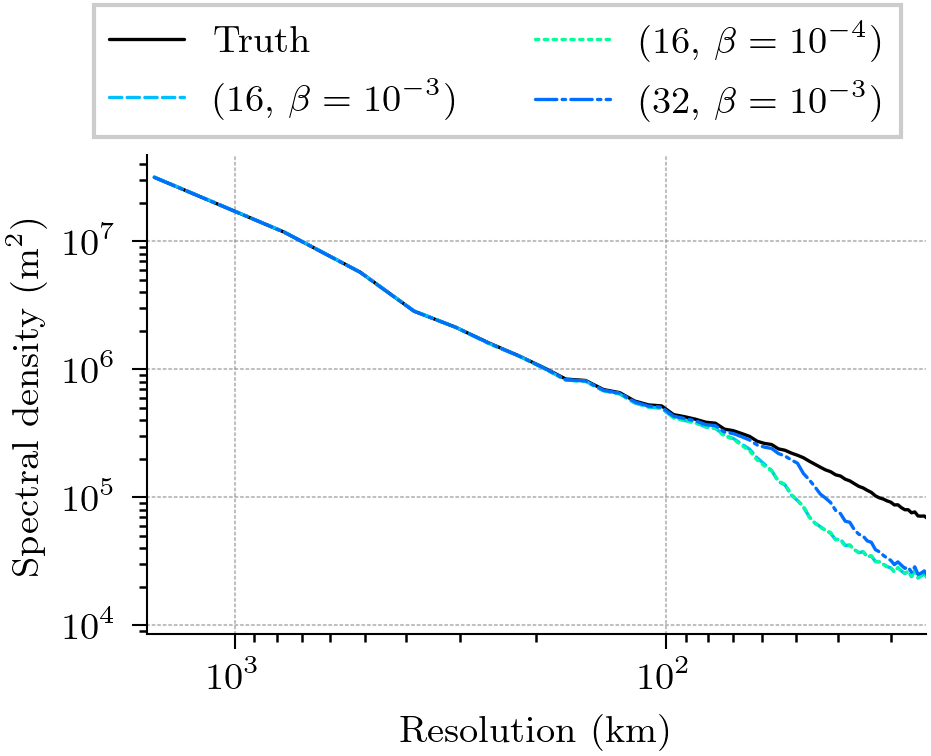}}
\caption{
    Estimated spectral density of the sea-ice thickness in the central Arctic for the testing dataset (black), and three different hyperparameter configurations for the autoencoder.
}
\label{fig:app_spectrum}
\end{center}
\end{figure}

In Fig. \ref{fig:app_spectrum}, we additionally show the energy spectrum for the sea-ice thickness if we increase the number of channels or lower the regularisation factor.
Once again, changing the regularisation has almost no impact on the averaged spectrum, while a larger latent space can seemingly retain more small-scale features.
Hence, if the regularisation is small enough, the compression rate determines how much small-scale features are retained through the latent space.
To reduce the smoothing, we can reduce the compression rate.
Since the smoothing is a result of double penalty effects caused by the point-wise comparison in the reconstruction loss, we however need other tools to completely remedy the smoothing.
Furthermore, by having more channels in latent space, the latent diffusion model can become more difficult to optimise \citep{rombach_high-resolution_2022}.

\subsection{Influence of the generated dataset size on the diffusion scores}\label{app:results_dataset_size}

To evaluate our diffusion models, we use $4383$ samples, while the recommendation to evaluate generative models for images is $50000$ samples.
In Table \ref{tab:app_results_dataset_size}, we compare the performance of the same latent diffusion model without censoring if we alter the random seed for the generation of the dataset.

\begin{table}[ht]
\caption{
    Evaluation of the statistics for samples generated by the latent diffusion model without censoring compared to the testing dataset with the same size.
}
\label{tab:app_results_dataset_size}
\begin{center}
\begin{small}
\begin{sc}
\begin{tabular}{lrr}
\toprule
Seed & $\mathrm{FAED}$ $\downarrow$ & $\text{RMSE}_{\text{SIE}}$ $\downarrow$ \\
\midrule
0 & 1.65 & 11.56 \\
10 & 1.66 & 11.79 \\
11 & 1.68 & 11.39 \\
12 & 1.66 & 11.76 \\
13 & 1.64 & 11.36 \\
\bottomrule
\end{tabular}
\end{sc}
\end{small}
\end{center}
\end{table}

While there are some small differences between different seeds, these differences are smaller than the differences shown in Tab. \ref{tab:diffusion}.
Based on this analysis, we can conclude that the dataset size is large enough to evaluate differences between diffusion models with the proposed scores.

Diffusion models tend to have a high volatility in the scores during training \cite{song_improved_2020}.
Normally used to stabilise the diffusion model, we restrain from applying exponential moving averages to estimate the scores of the diffusion model.
On the one hand, this allows us to establish the performance of diffusion models without such tricks.
On the other hand, the differences between the models for the $\mathrm{FAED}$ in Tab. \ref{tab:diffusion} might be due to the high volatility.

\subsection{Clipping of the generative diffusion model without censored distributions}\label{app:clipping}

To avoid unphysical samples during generation, the values from the denoiser are commonly clipped into either fixed \citep[e.g., ][]{ho_denoising_2020} or dynamical bounds \citep{saharia_image_2022}.
However, as shown in Appendix \ref{app:add_methods_diffusion}, generative diffusion is built around the assumption of unbounded Gaussian diffusion, and the clipping would violate the Gaussian assumption, which has been applied during training.
This could introduce biases into the generation.

In Table \ref{tab:app_clip}, we show the results of such a clipped diffusion model in addition to the results of the diffusion models as presented in the main manuscript.
For this clipped experiment, we use the pre-trained data-space diffusion model and clip the output of its denoiser into the physical bounds ($\text{SIT} \in [0, \infty)$, $\text{SIC} \in [0, 1]$, and $\text{SID} \in [0, 1]$).
Additionally, we show what happens if we clip the output of the latent diffusion model (LDM) trained without the censored distribution.

\begin{table}[ht]
\caption{
    Comparison of the diffusion models tested in Sec. \ref{sec:results} with diffusion model where the output of the denoiser is clipped into physical bounds or a latent diffusion model where the output of the autoencoder is clipped without a censored distribution.
}
\label{tab:app_clip}
\begin{center}
\begin{small}
\begin{sc}
\begin{tabular}{lrr}
\toprule
Model & $\mathrm{FAED}$ $\downarrow$ & $\text{RMSE}_{\text{SIE}}$ $\downarrow$ \\
\midrule
\textit{Validation}        & 2.38 & 7.02\\
\midrule
Diffusion                  & 1.73 & 9.66\\
Diffusion (denoiser clipped)        & 1.76 & 12.82\\
LDM                        & \textbf{1.65} & 11.56\\
LDM (censored)             & 1.79 & \textbf{7.32}\\
LDM (output clipped)              & 1.81 & 11.04\\

\bottomrule
\end{tabular}
\end{sc}
\end{small}
\end{center}
\end{table}

Including a clipping operator into the denoiser that defines the diffusion model in data space has a negative impact on the physical representation of the sea ice, the RMSE in the sea-ice extent is increased compared to the unclipped diffusion model.
While this technique is efficient in image generation, it introduces a bias into the denoiser, which seemingly hurts its performance in our sea-ice generator.
Because of biasing the diffusion model, we suspect that clipping introduces a degradation in the cross-correlation between variables, which can have a higher impact on geophysical applications than on image generation.

Clipping the output of a LDM trained with a Gaussian distribution has a neutral impact on the physical representation.
Since the model is trained without accounting for clipping during training, the model still produces spurious values for target values where the bounds are reached, only values exceeding the bounds are clipped.
For the targets that lay on the bounds, the censored distribution, introduced in Appendix \ref{app:censored_dist}, pushes the output of the LDM towards an extreme exceedance of the bounds, increasing the probability that the values is clipped into the bounds.
Hence, censored distributions can unlock an effective incorporation of physical bounds into the training of deep learning algorithms.  

This results suggests that we need special methods to properly treat physical bounds in diffusion models that work in data space.
We can define a log-barrier that pushes the diffusion process away from the bounds \citep{fishman_diffusion_2023}.
Additionally, we can define a reflected diffusion process, where the Brownian motion is reflected at the bounds \citep{lou_reflected_2023}.
However, with both methods, we would never generate values that exactly on the bounds.
If the result from the diffusion model is further used in downstream applications, e.g., by cycling in a surrogate model \citep{price_gencast_2024, finn_generative_2024}, such small errors can amplify and make the pipeline unstable.
Contrastingly, censoring of the autoencoder provides a clean and systematic way to incorporate bounds so that the generated data can lie exactly on the bounds.

\subsection{Censoring}

In Table \ref{tab:diffusion}, we show that censoring helps to improve the representation of the sea-ice extent and reduces the RMSE of the sea-ice extent.
Here, we disentangle a bit how we obtain this improvement.

\begin{figure}[ht]
\begin{center}
\centerline{\includegraphics[width=0.8\textwidth]{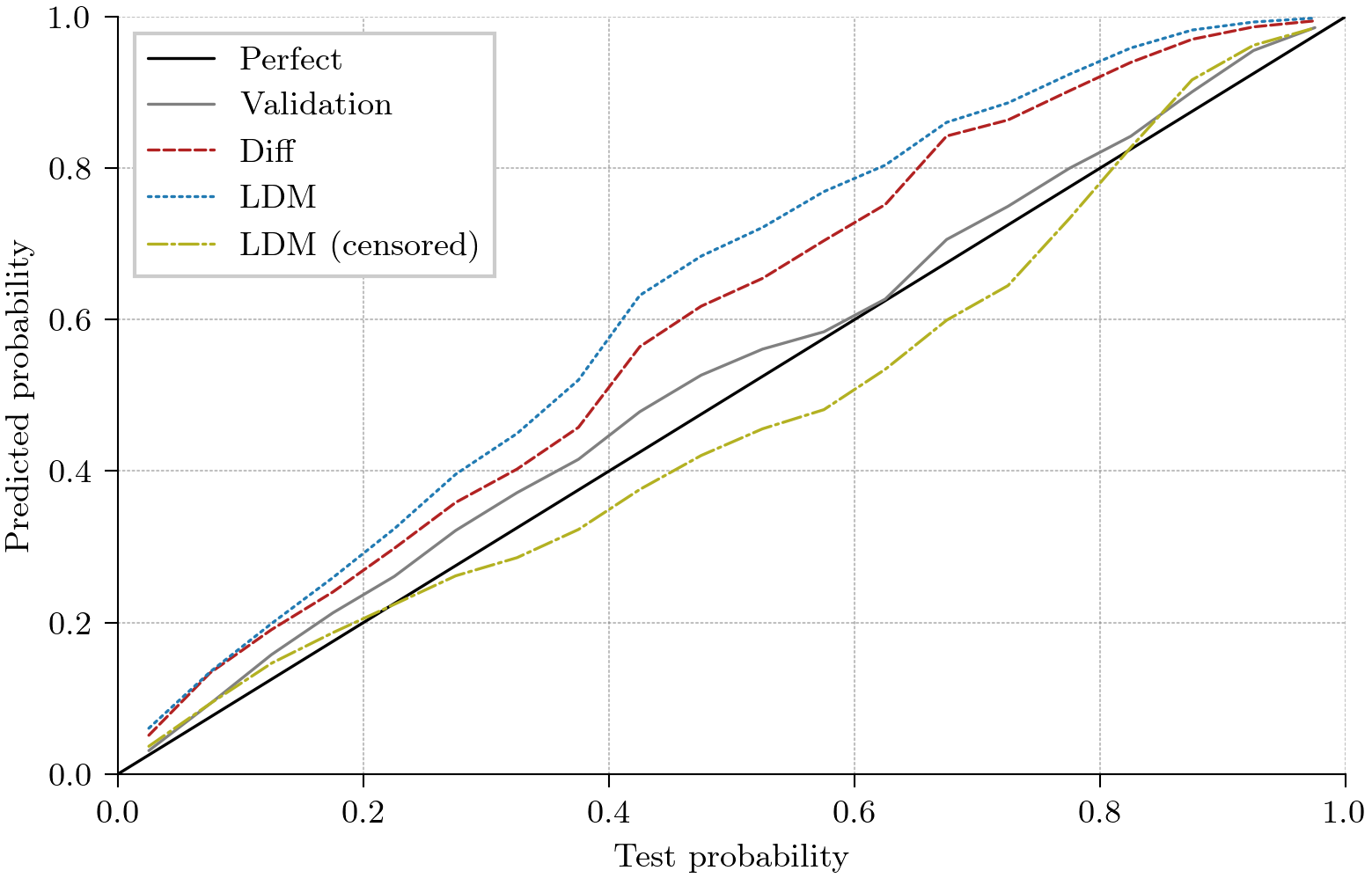}}
\caption{
    Comparison of the probabilities that a grid point is covered by sea-ice as seen in the testing dataset or as predicted by the validation dataset (gray), the diffusion model in data space (red, Diff), the latent diffusion model (blue, LDM), and the latent diffusion model with censoring (yellow).
    The shown probabilities represent the averaged predicted probabilities, grouped into $5\,\%$ intervals based on the test probabilities.
}
\label{fig:app_sie_prob}
\end{center}
\end{figure}

In Fig. \ref{fig:app_sie_prob}, we show the binarized probability that a grid point is covered by sea ice in the generated dataset dependent on the probability in the testing dataset.
The optimal probability would be on the one-to-one line, reaching for each grid point the same probability in the generated dataset.

In the validation dataset, the probabilities are slightly overestimated.
The difference is likely caused by the different dataset size (one year in validation and three years in testing).
As for the validation dataset, the diffusion model in data space and the latent diffusion model both overestimate the probabilities.
Since this overestimation is larger than for the validation dataset, it is likely because of the failure to represent the sea-ice extent correctly.
In contrast to the other diffusion models, the diffusion model with censoring tends to underestimate the probabilities.
The RMSE in Tab. \ref{tab:diffusion} is reduced as the underestimation for the censored model is smaller than the overestimation for the other diffusion models.
Consequently, we have established here that censoring helps to alleviate the overestimation bias of the diffusion models.
Its further use for other downstream tasks remains to be seen.

\subsection{Uncurated data samples}

\begin{figure}[ht]
\begin{center}
\centerline{\includegraphics[width=0.9\textwidth]{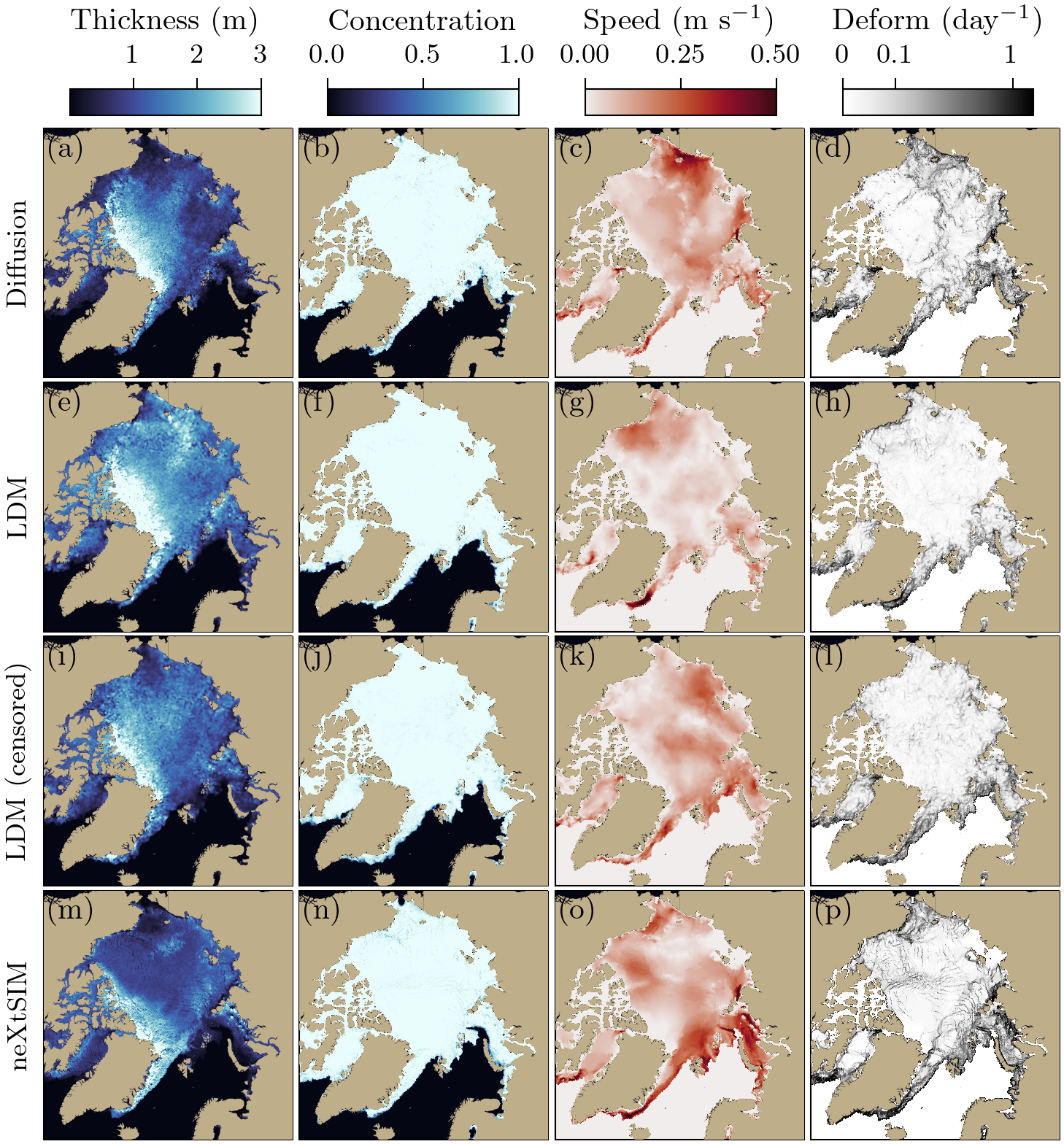}}
\caption{
    Uncurated samples from the diffusion model in data space (a--d), the latent diffusion model without censoring (e--h), and the latent diffusion model with censoring (i--l), and the neXtSIM simulations.
    As in Fig. \ref{fig:samples}, the thickness and concentration are directly generated, while the speed and deformation are derived from the generated velocities.
    Best to view in digital and colour.
}
\label{fig:app_uncurated}
\end{center}
\end{figure}

In Fig. \ref{fig:app_uncurated}, we show uncurated samples from our diffusion models and reference samples from neXtSIM.
As discussed in the main manuscript and shown in Fig. \ref{fig:samples}, the latent diffusion models result into smoothed fields compared to the diffusion model in data space and neXtSIM.

While the derived deformation field of the diffusion model (d) shows small-scale structures as they can be seen in neXtSIM (p), the thickness and concentration remain too noisy.
One possible explanation could be that the velocities are easier to generate as they are continuous, whereas the sea-ice thickness can be rather represented by discrete-continuous behaviour.
Since we used a rather low number of integration steps (20) without much tuning, we could expect that such fine-scale structure can be better represented with a better tuned sampling and/or with a diffusion model trained on more data.

\section{Changes to the Version 1}
\begin{itemize}
    \item We added the comparison in speed to Table \ref{tab:diffusion} indicating that latent diffusion models are indeed much faster than diffusion models in data space.
    \item We strengthened the proof-of-concept character of the study, indicating that we see this work as a first step towards the training of generative diffusion for e.g., Arctic-wide surrogate models.
    \item We added Appendix \ref{app:clipping} with additional results if we add clipping to the denoiser in data space. While in image generation, this clipping is very important, we find a negative impact on the physical representation in the diffusion model.
    We additionally provide insides why a censored distribution helps to improve the physical representation.
    \item We added the GitHub link to the code.
\end{itemize}

\end{document}